\documentclass[nonacm]{acmart}

\AtBeginDocument{%
  }

\usepackage{svg}
\setcopyright{none}
\usepackage{xcolor}
\usepackage{colortbl}
\begin{document}

\title{CoPE: A Small Language Model for Steerable and Scalable Content Labeling}

\author{Samidh Chakrabarti}
\authornote{Both authors contributed equally to this research. Author order determined by coin flip. $\dagger$ \textrm{Top contributor. Work completed while at Stanford.}}
\email{samidh@zentropi.ai}
\author{David Willner}
\authornotemark[1]
\email{dave@zentropi.ai}
\affiliation{%
  \institution{Zentropi}
  \city{San Francisco}
  \state{California}
  \country{USA}
}

\author{Kevin Klyman}
\authornotemark[2]
\affiliation{%
  \institution{Stanford University}
  \city{Stanford}
  \state{California}
  \country{USA}}

\affiliation{%
  \institution{Harvard University}
  \city{Cambridge}
  \state{Massachusets}
  \country{USA}
}

\author{Tiffany Saade}
\affiliation{%
  \institution{Stanford University}
  \city{Stanford}
  \state{California}
  \country{USA}}

\author{Emily Capstick}
\affiliation{%
  \institution{Stanford University}
  \city{Stanford}
  \state{California}
  \country{USA}}

\author{Sabina Nong}
\affiliation{%
  \institution{Stanford University}
  \city{Stanford}
  \state{California}
  \country{USA. Copyright to the authors, 2025.}}

\renewcommand{\shortauthors}{Chakrabarti and Willner et al.}

\begin{abstract}
  This paper details the methodology behind CoPE, a policy-steerable small language model capable of fast and accurate content labeling. 
We present a novel training curricula called \textit{Contradictory Example Training} that enables the model to learn policy interpretation rather than mere policy memorization. 
We also present a novel method for generating content policies, called \textit{Binocular Labeling}, which enables rapid construction of unambiguous training datasets. When evaluated across seven different harm areas, CoPE exhibits equal or superior accuracy to frontier models at only ~1\% of their size. We openly release a 9 billion parameter version of the model that can be run on a single consumer-grade GPU. 
Models like CoPE represent a paradigm shift for classifier systems. By turning an ML task into a policy writing task, CoPE opens up new design possibilities for the governance of online platforms.
\end{abstract}



\maketitle

\section{Introduction}
Content moderation at scale has proven to be one of the most intractable challenges of the digital age \citep{masnick2019impossibility}. It is only becoming more difficult as generative AI systems proliferate and user-generated content scales across languages, cultures, modalities, and platforms. Despite its difficulty, moderation cannot be avoided. To keep users safe and platforms usable in the face of spam, fraud, and abuse, it is necessary to enforce content policies informed by laws and ethical principles, all while protecting freedom of expression.

And yet, despite the fact that we are now twenty years into an era of wide spread, mass-scale content moderation ushered in by the rise of social media, traditional moderation systems simply do not work very well. Human content reviewers, who are the foundation of every existing enforcement system, are tightly constrained in their efficacy by both the individual challenges of cognitive and emotional overload, and the collective difficulty of coordinating human decisions across hundreds or even thousands of decision makers, undermining the accuracy, scalability, and fairness of enforcement decisions. Existing automated moderation tools, including rule-based filters and machine-learning classifiers, are respectively either insufficiently nuanced or difficult to update, and so struggle to keep pace with the incredible speed at which online content evolves. Content policies, which rely on predefined harm taxonomies, are similarly difficult to keep up to date with new types and presentations of harms as they emerge, and have trouble accommodating differences in values across languages, culture, and regions. 

The next generation of content moderation tools can overcome these challenges with new capabilities to respond to changes in policy, novel risks, and deliver high-fidelity judgments across diverse contexts without sacrificing efficiency or reliability. These tools will continue the trend of decreased reliance on mass-scale human data labeling, yielding not only more accurate results, but also requiring less intensive graphic exposure for human reviewers.  

In this context, transformer-based machine learning models theoretically present a promising alternative due to their ability to directly interpret natural language and process extended contexts. However, off-the-shelf LLMs are cumbersome to use in content moderation tasks \citep{willner2024policyDrivenTechPolicyPress}. These models face significant barriers to effective deployment for content moderation, including a lack of specialized training data, high inference costs, high latency, and bias.

We introduce CoPE (Content Policy Evaluator)—a specialized language model designed for automated, cost-effective, policy-agnostic, accurate, and rapid content policy evaluation.\footnote{The model is available on Hugging Face here \url{https://huggingface.co/zentropi-ai/cope-a-9b}} CoPE is built with the following key features:  
\begin{itemize}
    \item Pluralistic natural language content policies: The model allows developers to define and enforce diverse policy standards without relying on predefined value judgments.
    \item Reliable performance: The model demonstrates high accuracy and strong F1 scores across various harm areas including hate speech, sexual content, violent threats, harassment, self-harm, drug sales, as well as incivility/toxicity.
    \item Efficiency: The model operates with sub-200ms latency on an A100, ensuring fast response times.
    \item Local implementation: At 9 billion parameters, the model is small enough to be run locally without requiring a high-end GPU. 
    \item Open deployment: We release the model under an OpenRAIL license, allowing for broad use and adaptation by the trust and safety community. 
\end{itemize}

\section{Background and Related Work}
\subsection{Automated Content Moderation}
Early automated content moderation systems relied heavily on rule-based methods such as keyword filtering and hash-matching, later extending to machine-learning based approaches trained on patterns in human-labeled data. These approaches offer efficient mechanisms for detecting known violations, but lack the flexibility to handle nuance, new contexts, or subtle alterations \citep{engstrom2017limitsFiltering, reda2017whenFiltersFail}. Classifiers are often embedded in multistage moderation pipelines, particularly for larger social media platforms, where content is triaged based on confidence levels or severity scores from a classifier to enable differentiated downstream handling, ranging from automated action to escalation to human reviewers \citep{caplan2018contentOrContextModeration}. This structure has enabled platforms to scale content moderation to billions of users while preserving human oversight for the most difficult decisions.

However, traditional automated content moderation tools face several intractable challenges. First, they still require substantial human labor to create and maintain. \citet{roberts2019behindTheScreen} points out that platforms have preferred human content moderators based on considerations for nuanced interpretations, costs, and capacity. \citet{tubaro2020trainerVerifierImitator} emphasize that only a small portion of content moderation can be fully automated, since emerging types of data demand both new rounds of data labeling and model retraining by human workers. Similarly, \citet{jhaver2019automoderator} argue that automation does not eliminate human labor in content moderation but reshapes where it is needed, including monitoring of the system, reviewing appeals, explaining decisions, and adjusting policies. 

Second, these tools struggle to generalize to novel content patterns and cannot accommodate new policies at all without retraining. \citet{son2023contentModerationWild} note that platforms often train and deploy separate models for each moderation policy, which are divergent across host countries and underlying products. \citet{gomez2024algorithmicArbitrariness} further demonstrate that moderation classifiers of comparable accuracy can assign inconsistent labels for the same content, driven by seemingly minor choices during model development. 

\subsection{LLM-Based Approaches}
In response to the dependence on mass-scale human labeling and rigidity of earlier moderation systems, recent work has applied transformer-based language models for content moderation \citep{10.1609/aaai.v37i12.26752}. Unlike previous machine learning approaches, LLMs are more likely to understand nuanced content, perform natural language reasoning, and generalize across policy and harm contexts. 

A number of systems, such as OpenAI's Omni-Moderation, Google's ShieldGemma, and Ai2's WildGuard, adapted traditional classifier frameworks by fine-tuning LLMs for binary safety classification. However, their effectiveness remains constrained by fixed harm taxonomies—predetermined categories that require model retraining rather than simple policy document updates to accommodate new or modified content standards. BingoGuard extends this functionality by incorporating harm category identification, though still within a bounded taxonomy. More recent approaches including GPT OSS Safeguard 20b and  STAND-guard introduce greater adaptability through few-shot prompting or policy-as-prompt mechanisms, allowing models to generalize to the evolving moderation frameworks without retraining. Systems such as Guard Reasoner leverage the reasoning capabilities of recent LLMs to provide moderation rationales and chains-of-thought, offering more interpretable outputs to assist human decision making.  

\begin{table*}[t]
\centering
\small
\setlength{\tabcolsep}{2pt}
\begin{tabular}{p{2cm}p{2.5cm}p{1.3cm}p{3cm}p{3.5cm}p{2cm}}
\toprule
\textbf{Method} & \textbf{Base Model} & \textbf{Modality} & \textbf{Policy Adaptability} & \textbf{Output Granularity} & \textbf{Openness} \\
\midrule
BingoGuard \citep{yin2025bingoguard} &
Llama-3.1-8B &
Text &
Pre-defined policies &
Binary Classification + \newline Severity Scale &
Open weights \& fine-tuning data 
\\
\midrule
GPT OSS Safeguard 20b \citep{openai2025gptOssSafeguard} &
GPT-OSS &
Text &
Policy-as-prompt &
Multi-step reasoning chains + Classification &
Open weights \\
\midrule
Guard Reasoner \citep{liu2025guardreasoner} &
LLaMA 3.2 1B, 3B LLaMA 3.1 8B &
Text &
Limited to training-time \newline data synthesis &
Multi-step reasoning chains + Binary Classification &
Open weights \& fine-tuning data\\
\midrule
IBM Granite \newline Guardian 3.2 \citep{padhi-etal-2025-granite} &
Granite 3.2 3B-A800M Granite 3.1 8B &
Text &
Policy-as-prompt &
Classification + \newline Confidence Score &
Open weights \\
\midrule
Llama Guard 4 \newline \citep{inan2023llamaguardllmbasedinputoutput,chi2024llamaGuard3Vision} &
Llama 4 Scout &
Text \newline Vision &
Zero-/few-shot \newline promptable &
Binary Classification + \newline Violated Taxonomy &
Open weights \\
\midrule
Mistral Moderation API \citep{mistral2025guardrailingDocs} &
Ministral 8B 24.10 &
Text &
Pre-defined policies &
Binary Classification per harm category + Severity Scale &
API \\
\midrule
OpenAI omni- \newline moderation \citep{openai2025moderationDocs} &
GPT-4o &
Text \newline Vision &
Pre-defined policies &
Binary Classification per harm category + Confidence Score &
API \\
\midrule
SafetyAnalyst \citep{li2024safetyAnalyst} &
Llama 3.1-8B-Instruct &
Text &
Pre-defined policies &
Multi-label + reasoning &
Open weights \& fine-tuning data \\
\midrule
ShieldGemma \citep{zeng2024shieldgemma} &
Gemma 2 &
Text \newline Vision &
Pre-defined policies &
Binary Classification &
Open weights \\
\midrule
STAND-Guard \citep{wang2024standGuard} &
Mistral-7B-v0.1 &
Text &
Policy-as-prompt &
Binary Classification &
Not released 
\\
\midrule
WildGuard \citep{10.5555/3737916.3738177} &
Mistral-7B-v0.3 &
Text &
Pre-defined policies &
Binary Classification 
&
Open weights \& data \\
\bottomrule
\end{tabular}
\caption{\textbf{Comparison of Safety Moderation / Guardrail Methods.}}
\end{table*}

\subsection{Policy as Code}

In recent years, the cybersecurity community has increasingly adopted related methods, namely policy-as-code, where organizations write policies in a machine-readable format and allow them to be enforced by software systems \citep{ray2024policyAsCode}. This allows organizations to enforce privileges, access, and security controls at scale.

Several works have recently adopted this approach to improve the security of AI agents. \citet{shi2025progentprogrammableprivilegecontrol} develop “programmable privilege control for LLM agents,” which generates machine-readable security policies, restricts agents’ access to tools depending on the user request, and dynamically updates security policies based on the agent’s state. \citet{kholkar2025policyaspromptturningaigovernance} transform guardrail security policies into a verifiable policy tree, then analyze whether the agent is complying with these policies at runtime. Our approach develops analogous policy-as-code methods for scalable labeling of content. 

\subsection{Limitations}
Existing LLM-based approaches to content moderation do not adequately fulfill the demand for a system specialized for all-scenario content moderation, steerable to out-of-distribution harm taxonomies and policy guidelines, and open to community-driven evaluation, adaptation, improvement, and low-threshold adoption. As shown in the table above, few models offer unrestricted access to training and evaluation data, code, and model weights. The evaluation data landscape is fragmented: while some academic benchmarks are publicly available (e.g., Ethos), platforms rarely release the internal test sets they use to validate production systems. Furthermore, even systems with 'open' taxonomies often rely on proprietary severity scoring data to determine enforcement thresholds—data that shapes how policies are actually applied but remains undisclosed. This opacity makes independent evaluation difficult and limits reproducibility of reported results. Furthermore, among the major LLM-based content moderation systems we surveyed, with the exception of GPT OSS Safeguard, STAND-Guard and Granite Guardian, none support deliberate policy-as-input mechanisms that allow end users to steer model behavior according to evolving moderation guidelines. This limits their ability to generalize to emerging harm categories or platform-specific policies. 

Beyond adaptability constraints, existing LLM-based approaches face significant operational barriers. Frontier models like GPT-4o require substantial inference costs—often measured in cents per API call—making them economically prohibitive at platform scale where billions of moderation decisions occur daily. Latency presents an additional constraint: models exceeding 100ms response time cannot support real-time content filtering, forcing platforms to choose between post-publication moderation (which allows harmful content to circulate) or unacceptable user experience delays. Hardware requirements compound these challenges, as most state-of-the-art systems require high-end GPUs unavailable to smaller platforms or community moderators, limiting access to organizations with substantial computational resources.

In addition, some models are intended specifically as safety classifiers to add for user-facing LLMs (e.g., LlamaGuard, ShieldGemma, and Legilimens). As a result, may not be fit for purpose as general-purpose moderation frameworks capable of evaluating user-generated content across modalities and platforms.

The literature has also highlighted a number of remaining gaps of LLM-based approaches towards content moderation tasks.
\begin{itemize}
    \item \textbf{Prompting is not enough.} \citet{palla2025policyAsPrompt} argue that prompting language models directly with policies for moderation purposes leads to inconsistent outputs due to the model's sensitivity to minor variations in prompt phrasing. Similarly, \citep{gomez2024algorithmicArbitrariness} show that even fine-tuned LLMs for content moderation yield arbitrary and inconsistent decisions due to predictive multiplicity, where small training variations like random seed selection can lead to unstable moderation outcomes. 
    \item \textbf{Overfitting and model rigidity interfere with generalization and usability.} \citet{ma2024adaptinglargelanguagemodels} argue that LLM-based moderation systems often overfit to limited policy datasets, making them unreliable for unseen or evolving policy scenarios unless enhanced with techniques like chain-of-thought augmentation to improve generalization. \citet{kumar2025noFreeLunchGuardrails} argue that stronger guardrails in LLMs reduce harmful outputs but also degrade usability, implying a need for policy-agnostic models that can flexibly apply moderation without hard wiring fixed safety constraints. 
    \item \textbf{Lack of robustness and confidence calibration undermines reliability.} \citet{lu-etal-2025-llm} observe that LLM-based moderation models often fail to deliver reliable moderation results under distribution shifts or adversarial inputs, as their predictions tend to be overconfident.
\end{itemize}

\section{Objectives}
A policy interpretation model must read a written content policy, examine a piece of content, and determine whether that content violates the policy. There are three principal objectives for such a system: (i) steerability across diverse policy frameworks, (ii) accuracy in interpreting complex and nuanced policies, and (iii) efficiency for real-time, high-volume moderation. Each objective corresponds to a current market failure in the content moderation ecosystem. 

\subsection{Steerability}
Content policy frameworks differ substantially based on the type of platform for which the content policy applies, the modality of the data on that platform, the goals of the organization operating the platform, the level of (de)centralization of policy and moderation decision making, and the goal of any particular moderation task. Social media platforms differ from platforms for gig work or for online tutoring in important ways, necessitating different kinds of content policies (and different means of evaluating such policies). Some platforms are primarily text based (e.g., Reddit), whereas others are image-based (e.g., Pinterest, Instagram), video-based (e.g., YouTube, TikTok), or multimodal (e.g., Facebook), each of which require different considerations for content policies (e.g., child sexual abuse material in images). The organizations that operate platforms also have widely diverging goals, with some focused solely on maximizing profit and others oriented around building community, or creating space for certain kinds of discourse. 

While a profit-seeking platform owner might prioritize restricting content that would antagonize advertisers, a non-profit, community-oriented platform owner might prioritize restricting content that makes communities on the platform feel unwelcome. Some platforms (e.g. Facebook, Youtube, TikTok) operate highly centralized moderation systems, where uniform policies are set and controlled by the platform owner, where others (e.g. Reddit, Mastodon, Bluesky) operate systems that devolve control to community members in various ways, resulting in a much more diverse array of standards. In addition, teams within platforms frequently have the need to prioritize specific kinds of content, or specific subsets of it, resulting in further diversity in the frameworks being applied. 

Machine learning models that evaluate content policies must be steerable across diverse policy frameworks to be general-purpose. Non-modular evaluation pipelines serve only specific policy subsets and become obsolete as platforms evolve. Non-steerable systems take two forms, both inadequate. Rule-based systems rely on hard-coded lists of violative patterns (keyword filters, hash databases), which cannot accommodate context-dependent judgments. Traditional ML classifiers train on fixed taxonomies of harm categories, requiring extensive labeled data and costly retraining to accommodate policy changes. Neither approach serves platforms with diverse or evolving standards, as content deemed violative under one policy may be explicitly permitted or even promoted under another. 

In our Training Curricula Methodology section, we describe how we address this problem by carefully constructing a training set through \textbf{Contradictory Example Training}. 

\subsection{Fidelity}
Faithful interpretations of the plain meaning of complex and nuanced content policies is another essential desiderata of model-based evaluation. While human escalation and content policy teams can account for edge cases by using their broader understanding of platform-owner priorities, the slowness of that process means they can only feasibly do so for a tiny fraction of all moderation decisions. The scale and repetitiveness of online content means that such careful examination is only actually worthwhile in a vanishingly small fraction of cases. As a result, the vast majority of all moderation decisions on any large platform are made by a combination of frontline human reviewers, who are largely not \textit{allowed} to use discretion, and machine learning models, which are incapable of doing so. 

To ensure predictable and consistent interpretation between different human reviewers (and therefore by the machine learning models trained to predict what  their decisions would be for any given piece of content), the content policies written for application by frontline teams are generally intended to be black and white codes, designed to be followed with as little application of any specific person’s judgement as possible. This often results in standards that are quite exacting, as policy writing teams attempt to account for an ever-growing number of nuances and edge cases, and frequently include instructions on when to escalate content for more careful or detailed review. 

Large language models for policy interpretation must accurately parse detailed standards and provide confidence estimates grounded in specific policy provisions. Without this reliability, deployment teams cannot predict model behavior in response to policy instructions, rendering the system unfit for production use.

In our Policy Generation Methodology section, we further describe how we ensure a high volume of reliable labels in our training set by scaling data generation with an approach we call \textbf{Binocular Labeling}.

\subsection{Efficiency}
Efficiency is a sine qua non of any useful content moderation system for a set of overlapping reasons. The best known of these is the sheer volume of content requiring moderation.  Almost all online platforms are volume businesses -- while they may make enormous amounts of money in aggregate, they make very little on average \textit{per-piece-of-content}. This means that, almost irrespective of the overall size of a platform, the number of decisions moderation teams must make is enormous when compared to the people, hardware, and money available to them. This challenge of resource-relative-scale is only becoming more severe with the proliferation of generative AI, which can mass produce even more content, requiring additional moderation automation. 

Less well known, but equally important, are the interface latency expectations of platform users. In an era of increasingly fast computers and internet connections, people expect the systems they use to feel responsive and work more or less instantly. Since automated moderation systems have to run on every piece of content to detect violations, they must either run very quickly (limiting accuracy) or run after content has been posted (limiting the ability to prevent harm).

For machine-learning based systems specifically, there is yet another factor in the current bottleneck of appropriate hardware available for inference inside of any given platform. While it is reasonable to wish it were otherwise, the teams responsible for moderation systems rarely get the first pick of limited resources. This reflects a structural tension in for-profit platforms: content moderation is operationally essential but generates no direct revenue, creating persistent pressure to minimize costs even as moderation demands grow. As a result, any system like the one we propose becomes dramatically more useful in practice the more efficient it is, in terms of the related variables of cost, speed, and required hardware. This goal guided our selection of a relatively small base model (Gemma 2 9b) on which to experiment \citep{gemmateam2024gemma2improvingopen}.

\section{Training Curricula Methodology}

To achieve our objectives of steerability and fidelity, we developed a novel training curricula centered on what we have termed \textit{Contradictory Example Training}. We were able to train a model to follow clearly unambiguous policy instructions with a high degree of precision by exposing it to a dataset containing numerous policy-example-label triplets, each of which is internally consistent but contradicts others due to differences across policies. We defined determinism as the degree to which independent readers could use the policy to achieve a high inter-rater reliability (that is, could reach identical conclusions when applying a policy to the same content samples without conferring). This determinism is critical: without it, the model could succeed through guesswork or cultural intuition rather than by actually interpreting policy text. Put another way, this emphasis on parsability and consistency guided our policy design principles throughout the data generation process.

This approach ensures that the model cannot succeed by learning generalized patterns or cultural heuristics about content moderation. Our training dataset includes policies spanning the full spectrum of positions within each harm area—from highly permissive to extremely restrictive. By training on these diverse contradictory policies, the model learns to generalize to any intermediate position. For example, if the training data includes both a broad hate speech policy that prohibits microaggressions across many protected categories and a narrow policy that prohibits only violent threats against a few specific groups, the model learns to faithfully implement any policy formulation between these extremes—without requiring explicit training on every possible variation..

The effectiveness of this methodology hinges on the unambiguous nature of the policies employed. Each policy must, in so far as it is possible given the inherent limitations of language, dictate a clear and unambiguous classification for any given content example, establishing a single correct answer that can serve as ground truth during training. The specific nature of these answers—whether they represent wise or practical moderation decisions—is immaterial to the training process. What matters is the clarity and precision with which the policy prescribes outcomes and the diversity of contradictory stances among policies available for training.

By constructing a dataset that presents the model with numerous clearly unambiguous policies yielding contradictory outcomes for similar content, we create a training environment where success is only achievable through strict adherence to the explicit instructions provided for each example. This stands in sharp contrast to traditional moderation models, which learn through pattern matching on cultural heuristics—identifying content that 'looks like' hate speech based on word associations or demographic mentions. Such models may correctly classify obvious examples but struggle with edge cases where cultural context differs from training distributions. By making policy adherence the only viable strategy through contradictory examples, we eliminate the model's ability to rely on these shortcuts. The model cannot rely on learning only the gist of our policy perspectives, as the dataset deliberately precludes the internalization of any coherent set of values or the application of general cultural expectations, therefore avoiding the presentation of a consistent conceptual framework. Instead, the only viable path to accurate performance requires the model to focus on the precise textual content of the specific instructions provided for each example.

\begin{figure}[t]
\centering
\includegraphics[width=\columnwidth]{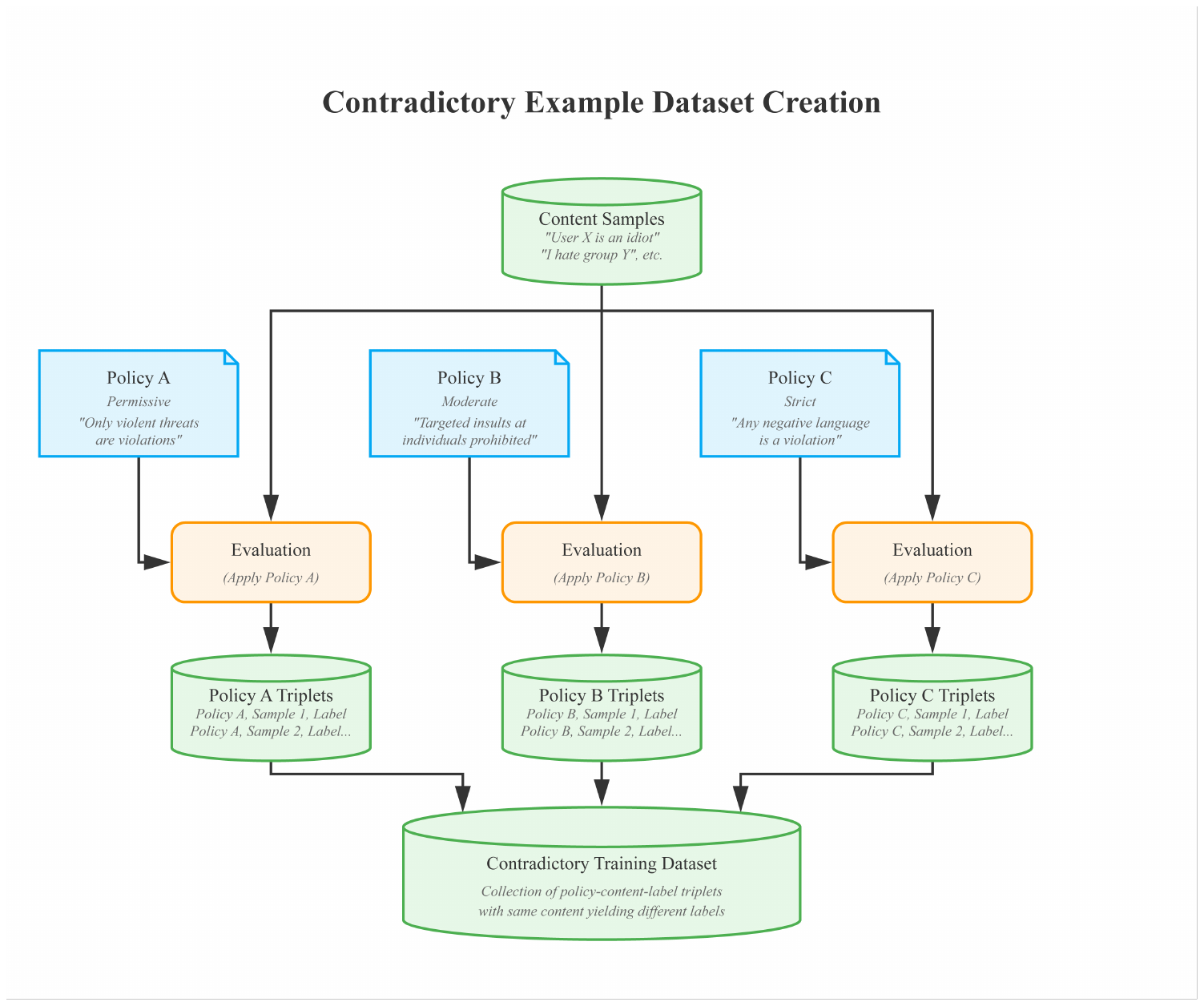}
\caption{Training Curricula Creation Process via Contradictory Examples}
\label{fig:yourlabel}
\end{figure}

\section{Policy Generation Methodology}

\subsection{Format and Subject Matter Selection}
The implementation of our training methodology critically relied on a well developed understanding of what constitutes a “clearly unambiguous” policy, as defined in the section above. Our policies follow a standardized format: each consists of a clear definition of the harm area, a specification of what content violates the policy (the inclusion criteria), a list of exceptions or edge cases (the exclusion criteria), and examples demonstrating borderline cases. To ensure the requisite level of determinism, we imposed several strict constraints on our policy construction. 
\begin{itemize}
    \item \textbf{Controlled Experimental Design:} We sought to minimize extraneous variables in our experimental design, and so restricted our experimentation to text content (as opposed to other modalities) with English-language policies and content samples. 
    \item \textbf{Binary Classification:} In order to ensure that data labeled according to each policy was directly comparable across both competing definitions of a given issue and across different issue areas, we constrained our policies to binary classification only.
    \item \textbf{Standardized Semantic Contrast:} We prioritized contrasts in semantic meaning over variations in diction, ensuring that the model would learn to attend to substantive policy differences rather than superficial stylistic variations. This led us to establish a standardized policy format across all severities and harm types.
    \item \textbf{Explicit Semantic Evaluation:} We restricted policies to addressing information explicitly present in the policy. We deliberately avoided including criteria that would require additional information or subjective interpretation, such as assessing the truthfulness of claims or inferring the intentions of authors.
    \item \textbf{Focused Policy Domains:} We consider policy domains that are centered on the assessment of semantic content—areas such as hate speech detection, where evaluation depends on textual meaning rather than external verification. 
    \item \textbf{Spectrum of Intensity:} We selected subject matter areas where we could find data spanning a broad spectrum of ``intensity'' within the area, as this diversity allowed us to evaluate adherence to both strict and permissive policy variations. In practice this meant working across hate speech, sexual content, violent threats, harassment, self-harm, drug sales, and toxicity (narrowly defined as incivility directed towards other conversation participants to differentiate it from harassment).
\end{itemize}

The main limitation is that contradictory training requires deterministic policies—those that achieve $\geq$0.9 F1 inter-rater agreement among humans. Highly subjective categories may not meet this threshold—though this constraint may itself be valuable, as it forces explicit articulation of moderation criteria rather than reliance on assumed shared judgment.

\subsection{Policy Refinement}
The contradictory training approach depends entirely on policy clarity. If policies are ambiguous, the model can succeed by learning general cultural heuristics about what content `seems' problematic rather than learning to follow specific written instructions. For example, a vague hate speech policy like ``prohibit offensive content about protected groups'' allows the model to rely on broad cultural associations about offensiveness. In contrast, a deterministic policy like ``prohibit content that calls for violence [defined as follows] against members of [specific list of groups]'' forces the model to parse the actual policy text, particularly for policies with atypical definitions of violence or unusual lists of protected groups.

By ensuring each policy produces consistent labels ($\geq$0.9 inter-rater agreement), we guarantee that successful performance requires reading and applying the policy rather than pattern matching. When policies are unambiguous but contradictory - where identical content receives different labels under different policies - the only path to accuracy is strict policy adherence.

To create the clearly unambiguous policies needed for our training method to work, we developed a \textit{Binocular Labeling} methodology that uses AI-generated labels as a diagnostic tool for policy improvement. By analyzing how an LLM-powered labeling system (SafetyKit) interpreted various phrasings of the same concept, we were able to systematically identify and resolve ambiguities in policy language -- specifically, words or phrases that led to inconsistent interpretations across rephrasings, edge cases that produced different labels under semantically equivalent policies, and definitional gaps that left room for subjective judgment -- instead of relying solely on human judgments of policy clarity. This approach transformed the challenge of creating clear policies from a purely subjective task into an empirically-guided one, increasing the speed and rigor of both our policy drafting and labeling. The process combined automated labeling using SafetyKit (a multi-pass chain-of-thought moderation tool powered by GPT-4 that can take custom text-based policies as input) with strategic manual review as follows:

First, we manually drafted an initial content policy (P$_1$-Main) and then entered that policy into a large language model (in this case Claude 3 Opus) with the prompt “Please rewrite this document, changing the language as much as you can while keeping the style, format, length, and meaning identical.” This procedure produced a linguistically distinct, but semantically very similar, alternative articulation of that policy (P$_1$-Alt). We then lightly manually corrected P$_1$-Alt to fix formatting issues, simplify sentences where the LLM used overly complex language, and clean up obscure or strange word choices. 

After finalizing P$_1$-Alt we used it to process unlabeled content samples from open source data repositories through SafetyKit. This produced a dataset (D$_1$-Alt) labeled according to the alternative policy articulation (P$_1$-Alt). We then reprocessed that dataset (D$_1$-Alt) through SafetyKit using the initial manually written policy (P$_1$-Main), resulting in a second labeled dataset (D$_1$-Main). The team then cross-referenced the labels in D$_1$-Main with those in D$_1$-Alt, set aside all content samples where the labels in the two data sets matched, and manually reviewed only the remaining mismatched content samples. When the label from D$_1$-Main presented a more faithful reading of P$_1$-Main than the label from D$_1$-Alt, they changed the label in D$_1$-Alt to match the label in D$_1$-Main. When the label from D$_1$-Alt presented a more faithful reading of P$_1$-Main than the label from D$_1$-Main, they kept the label in D$_1$-Alt. This approach allowed us to focus on cases where SafetyKit-generated labels differed notably due to relatively minor differences in policy wording, particularly emphasizing ambiguous edge cases and subtle interpretive variations, while avoiding spending time on very clear cut cases that received the same verdict under both P$_1$-Main and P$_1$-Alt.

Based on observations from the manual review process, and LLM-generated explanations SafetKit generates for every label, we then revised P$_1$-Main to clarify its meaning and account for edge cases, with the goal of creating a text that SafetyKit could more accurately interpret, producing a new draft (P$_2$-Main). Then we repeated the entire data labeling process: generating a (P$_2$-Alt) from (P$_2$-Main) using an LLM; labeling the unlabeled content samples using (P$_2$-Alt); relabeled that data using (P$_2$-Main); and manually reviewing the new set of mismatches. This process was then repeated until the F1 of (D$_N$-Main) compared to (D$_N$-Alt) was above 0.9, at which point (P$_N$-Main) was adopted as the final policy and (D$_N$-Main) served as the basis for ground-truth labeling.

After the final SafetyKit iteration, to ensure label quality and consistency, our team employed a disagreement-based manual correction approach to inspect discrepancies in the datasets labeled by SafetyKit under different policy iterations. Human labelers applied their expert discretion to reconcile edge cases and resolve ambiguities, finalizing the “ground truth” labels used for our training data.

While still quite time consuming, this process allowed us to avoid the need for human labeling of the vast majority of samples in each subject matter area, as well as the need for relabeling the entire data set every time we significantly altered a policy, saving a very significant amount of time compared to any similarly intensive multi-pass manual labeling process.

\begin{figure}[t]
\centering
\includegraphics[width=.9\columnwidth]{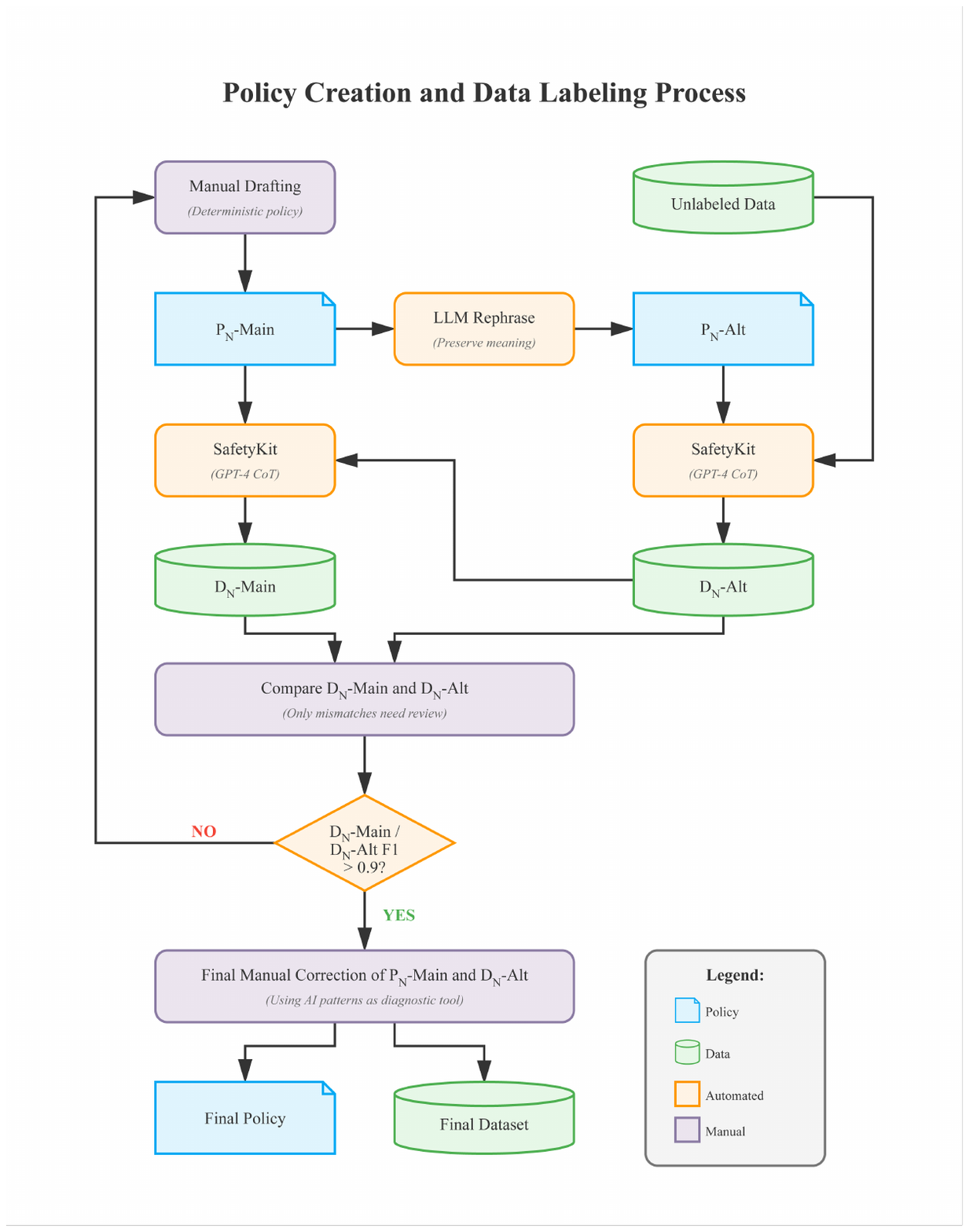}
\caption{Policy Generation Process via Binocular Labeling}
\label{fig:yourlabel}
\end{figure}
\clearpage

\subsection{Policy Variations}
Using the data labeling approach outlined above, we developed unambiguous "seed policies" and corresponding well-labeled datasets across all seven subject areas. These seed policies were designed to be “middle of the road” in terms of strictness, because moderate policies and nuanced policies typically contain both a large number of inclusions and a significant number of caveats and exclusions. We then systematically expanded each seed policy through a two-phase process designed to create contradictory training examples while testing different aspects of policy interpretation.\footnote{For examples of the types of policies we generated, see \url{https://zentropi.ai/u/dave}}

\textbf{1. Major Variations (Strictness + Linguistic Distinctness)} \newline
For each seed policy, we created four strictness variations spanning a spectrum from highly permissive to extremely stringent. Permissive policies were created by removing or narrowing inclusion categories and adding or expanding exception categories. Conversely, stringent policies were created by adding or expanding inclusion categories and removing or narrowing exclusions to capture a wider range of violations.
We then put these variations through the Policy Refinement and Data Labeling Methodology described above but, on the last iteration of the process, we reversed the roles of the Main and Alternate versions of the policies and datasets, adopting P$_N$-Alt as our final policy draft and D$_N$-Main as our final provisional dataset going into the Final Manual Correction step. This helped to ensure that each Variation was linguistically distinct from the seed policy and from every other Variation (e.g. each used different phrasing and, even when explaining similar concepts). The process resulted in the creation of five unique policies and corresponding datasets (the seed policy + 4 major variations) with systematically different ground truth labels based on policy strictness.

\textbf{2: Minor Variations (Definitional Subtleties)}
\newline For a subset of the subject matter areas, we then used each of the five major variations as starting points, we created 3-5 definitional variations per major policy by selectively removing specific subcategories or definitions (e.g., removing protected categories A and B while retaining C). Importantly, these definitional variations were not rephrased by the LLM, preserving subtle textual differences to test the model's sensitivity to minor policy modifications.
Each definitional variation was then used to relabel the same original content through SafetyKit, and the differences manually reviewed, creating additional contradictory labeled datasets where identical content receives different classification outcomes based solely on minor textual changes with significant definitional impacts (e.g., moving a specific class from protected to not protected in a hate speech policy).
This process transformed each seed policy into approximately 15-25 policy variations (5 major × 3-5 minor variations each), with each variation producing its own labeled dataset from the same underlying content. All of this data (arranged as policy-text, example, label triplets) across all subject matter areas were then merged to create our final training dataset. The result was a training corpus where identical content examples had systematically contradictory ground truth labels depending on the specific policy applied, making it impossible for the model to succeed through pattern memorization or cultural heuristics. Instead, the model needed to learn to follow the precise instructions provided in each policy to achieve accurate performance.

\section{Model Evaluation}

Using the methodology described above, we finetuned a LoRA adapter on top of Google’s Gemma 2 9B base model. We call this initial release CoPE-A-9b. The model accepts an arbitrary content policy and text content sample and outputs a single token (0 or 1) to indicate if the content adheres to the given policy. The model is released under an open license so that others may replicate the evaluation results we are sharing below.

\subsection{Challenges}
Evaluating a policy interpretation model presents unique challenges compared to traditional content classifiers. While conventional models can be assessed against fixed ground truth labels, CoPE's performance must be measured across varying policy definitions where the same content legitimately receives different labels. Furthermore, most publicly available benchmarks do not disclose the labeling guidelines that were given to raters, or the policies themselves are inadequately specified (e.g., ``label hate speech'' rather than defining what exactly constitutes hate speech). We therefore needed to create a novel evaluation framework to address these requirements.

\subsection{Framework}
To assess the performance of CoPE (and similar models), one must evaluate against policy-conditioned ground truth, where correctness depends both on the content samples as well as the specific policy applied. One cannot simply assume there is a single "correct" label for each content sample. Both the policy and the content sample must be fed to the model.
To do this properly and ensure the model cannot memorize policy-to-outcome mappings, we reserved a random set of content samples that were not seen during training as well as a random set of policy variations that were not used for training. These formed the test evaluation set, which was further formatted to match the prompt syntax that each model natively expects.
In total, we evaluated across seven distinct harm categories (hate speech, sexual content, self-harm, harassment, toxicity, violent threats, and drug sales) to ensure generalization beyond training distributions. We utilized third-party datasets for hate speech and toxicity to test outside the model's priors. Third-party datasets with disclosed labeling guidelines do not exist for the other harm categories we evaluated. 

\subsection{Baselines}

We evaluated CoPE relative to two categories of peer models. First, flexible policy models like GPT-4o that can accept arbitrary policies as prompts. Second, fixed taxonomy models such as LlamaGuard3-8B and ShieldGemma-9B, which excel within their trained definitions but cannot easily adapt to policies outside their training scope.

Given the heterogeneous nature of baseline models, we developed a fair comparison protocol. Where possible, all models were evaluated on identical content-policy pairs to ensure a common evaluation set. For fixed-taxonomy models, we evaluated only on policies within their scope through taxonomy mapping.

\subsection{Metrics}
We employed standard binary classification metrics adapted for policy-conditioned evaluation: Precision, recall, and F1 score (which computes a harmonic mean to account for imbalanced datasets). For each metric, we report both per-policy scores and aggregated scores across policy variations to assess consistency. While the base model supports multiple languages, we only tested on US English and did not yet assess multilingual performance.

\begin{table}[t]
\centering
\small
\begin{tabular}{lccc}
\toprule
\textbf{Model} & \textbf{Precision} & \textbf{Recall} & \textbf{F1 Score} \\
\midrule
CoPE-A-9B & 89\% & 93\% & \cellcolor{green!30}91\% \\
GPT-4o & 97\% & 78\% & 87\% \\
Llama-3.1-8B & 59\% & 96\% & 73\% \\
LlamaGuard3-8B & 88\% & 64\% & 74\% \\
ShieldGemma-9B & 68\% & 98\% & 80\% \\
\bottomrule
\end{tabular}
\caption{\textbf{Hate Speech Classification (Internal Set)}}
\label{tab:hate-speech-internal}
\end{table}

\begin{table}[t]
\centering
\small
\begin{tabular}{lccc}
\toprule
\textbf{Model} & \textbf{Precision} & \textbf{Recall} & \textbf{F1 Score} \\
\midrule
CoPE-A-9B & 80\% & 88\% & \cellcolor{green!30}84\% \\
GPT-4o & 91\% & 78\% & \cellcolor{green!30}84\% \\
Llama-3.1-8B & 62\% & 92\% & 74\% \\
LlamaGuard3-8B & 87\% & 79\% & 83\% \\
ShieldGemma-9B & 82\% & 82\% & 82\% \\
\bottomrule
\end{tabular}
\caption{\textbf{Hate Speech Ethos Benchmark \citep{mollas2022ethos}}}
\label{tab:hate-speech-ethos}
\end{table}

\begin{table}[t]
\centering
\small
\begin{tabular}{lccc}
\toprule
\textbf{Model} & \textbf{Precision} & \textbf{Recall} & \textbf{F1 Score} \\
\midrule
CoPE-A-9B & 93\% & 87\% & \cellcolor{green!30}90\% \\
GPT-4o & 64\% & 89\% & 75\% \\
Llama-3.1-8B & 33\% & 96\% & 49\% \\
\bottomrule
\end{tabular}
\caption{\textbf{Toxic Speech Classification (Internal Set)}}
\label{tab:toxic-speech}
\end{table}

\begin{table}[t]
\centering
\small
\begin{tabular}{lccc}
\toprule
\textbf{Model} & \textbf{Precision} & \textbf{Recall} & \textbf{F1 Score} \\
\midrule
CoPE-A-9B & 96\% & 83\% & \cellcolor{green!30}89\% \\
GPT-4o & 95\% & 72\% & 82\% \\
Llama-3.1-8B & 48\% & 95\% & 64\% \\
LlamaGuard3-8B & 100\% & 43\% & 60\% \\
ShieldGemma-9B & 96\% & 76\% & 85\% \\
\bottomrule
\end{tabular}
\caption{\textbf{Sexual Content Classification (Internal Set)}}
\label{tab:sexual-content}
\end{table}

\begin{table}[t]
\centering
\small
\begin{tabular}{lccc}
\toprule
\textbf{Model} & \textbf{Precision} & \textbf{Recall} & \textbf{F1 Score} \\
\midrule
CoPE-A-9B & 83\% & 93\% & \cellcolor{green!30}88\% \\
GPT-4o & 84\% & 93\% & \cellcolor{green!30}88\% \\
Llama-3.1-8B & 56\% & 96\% & 70\% \\
LlamaGuard3-8B & 65\% & 84\% & 73\% \\
ShieldGemma-9B & 69\% & 89\% & 78\% \\
\bottomrule
\end{tabular}
\caption{\textbf{Self-Harm Content Classification (Internal Set)}}
\label{tab:self-harm}
\end{table}

\begin{table}[t]
\centering
\small
\begin{tabular}{lccc}
\toprule
\textbf{Model} & \textbf{Precision} & \textbf{Recall} & \textbf{F1 Score} \\
\midrule
CoPE-A-9B & 69\% & 78\% & \cellcolor{green!30}73\% \\
GPT-4o & 100\% & 17\% & 30\% \\
Llama-3.1-8B & 35\% & 87\% & 50\% \\
ShieldGemma-9B & 49\% & 55\% & 52\% \\
\bottomrule
\end{tabular}
\caption{\textbf{Harassment Classification (Internal Set)}}
\label{tab:harassment}
\end{table}

\subsection{Results}
Our results demonstrate that CoPE-A achieves state-of-the-art performance in policy-based content classification, matching or exceeding GPT-4o while operating at a fraction of the size and computational cost of frontier models.
CoPE-A demonstrated superior performance in hate speech detection, achieving 91\% F1 score compared to GPT-4o's 87\%. Notably, CoPE-A maintained high precision (89\%) while achieving exceptional recall (93\%), indicating robust detection without excessive false positives. The Ethos benchmark provided external validation, where CoPE-A achieved 84\% F1, matching GPT-4o despite the benchmark's different policy formulation than our training data. This suggests strong generalization to unseen policy specifications.

Across all categories, CoPE-A exhibited more balanced precision-recall trade-offs compared to baselines. GPT-4o tends toward high precision (91-100\%) but variable recall (17-93\%), suggesting conservative classification. In contrast, Llama-3.1-8B shows high recall (87-96\%) but poor precision (33-62\%), indicating over-classification. CoPE-A maintains balanced performance with both metrics typically above 80\%, demonstrating neither excessive false positives nor missed violations.
Note that the fixed-taxonomy models (ShieldGemma and LlamaGuard) are omitted from tables where the harm area is outside their taxonomy. While we did calculate numbers for these models, they unsurprisingly performed sufficiently poorly that it seemed unfair to include them in the data above. This further highlights a fundamental limitation of traditional approaches.
Finally, it is also worth observing that CoPE-A’s performance on harassment was worse (F1 = 0.73) than other harm areas. Whether this reflects dataset inconsistencies or inherent difficulty remains unclear. This points to interesting future questions about which policy domains benefit most from contradictory training approaches. 

\subsection{Efficiency}
CoPE-A's architecture enables production-viable deployment across all critical dimensions. At approximately 1\% the size of frontier models, it is orders of magnitude faster and cheaper to run, which is required for the real-time labeling needs of internet scale platforms. CoPE-A can be run on a single consumer-grade GPU, such as an NVIDIA L40S, or even on edge devices.

\section{Discussion}

Our results validate the central premise that contradictory example training can produce models capable of true policy interpretation rather than policy memorization. This has potentially major implications for operational agility, policy development, and online governance.

\subsection{Implications for Operational Agility}
CoPE’s technical achievements address a fundamental barrier to effective content moderation: the inability to rapidly adapt classification systems to evolving policies without onerous retraining infrastructure. By decoupling policy from model training, organizations can modify content standards by editing policy documents rather than collecting new training data and retraining classifiers. This transforms content moderation from a machine learning problem requiring specialized expertise into a policy writing problem—a skill far more widely distributed among trust and safety professionals. Historically, policy changes have outpaced the ability to update ML-based moderation systems. This no longer needs to be the case.

\subsection{Implications for Policy Development}
The implications extend beyond operational efficiency. First, policy experimentation becomes feasible: trust and safety teams can test different policy formulations empirically rather than committing to approaches they cannot easily revise. Second, geographic and cultural adaptation can better scale: platforms can deploy region-specific policies without maintaining separate classification infrastructure for each locale. Third, policy pluralism becomes viable: different communities can enforce legitimately different standards using shared infrastructure, enabling values-based differentiation that was previously constrained by technical limitations.

This paradigm shift does not eliminate difficult normative questions about what content should be permitted—if anything, it makes those questions more salient by removing technical constraints as an excuse for inaction or inconsistency.

\subsection{Implications for Technical Research}
This work also suggests that the field may benefit from reconceptualizing automated enforcement as a policy interpretation problem rather than a classification problem. Much existing research treats moderation as requiring fixed taxonomies of harm, optimizing for accuracy within predefined categories. Our approach demonstrates that models can instead learn to parse policy language itself, enabling adaptation to evolving standards.

This shift has implications for how one approaches both the development and evaluation of moderation systems. Instead of asking "How accurate is this classifier?" it becomes more important to ask "How steerable is this classifier?" and “How clear is this policy?” At present, there are no well defined metrics for either of these concepts nor any benchmarks to do such comparative evaluation. We hope other researchers take up the mantle on these questions.

\subsection{Implications for Online Governance}
Finally, democratizing access to sophisticated moderation tools through smaller, steerable models may enable new forms of online governance. Community-run platforms, decentralized networks, and special-purpose spaces could enforce nuanced content policies tailored to their specific contexts and values. People can truly become participatory citizens of their online communities rather than merely users. Will this enable greater diversity in online speech (as we hope) or new vectors for abuse? These are questions beyond the scope of our technical research but essential to address as the technology matures.

\section{Ethical Considerations}

The capabilities that make CoPE useful for content moderation—steerability, accuracy, and efficiency—also create potential for misuse. This section considers those risks and the design decisions made to mitigate them.

\subsection{Dual-Use Risks}
CoPE's ability to enforce arbitrary policies creates obvious risks. Authoritarian governments already use machine learning for content filtering and surveillance \citep{doi:10.1126/sciadv.abl8198}. This includes but is not limited to: information suppression at key political moments, automated profiling and exclusion of activists, and stifling of political speech \citep{unver2024aiHumanRights}. \citet{Funk2023Repressive} report that “legal frameworks in at least 21 countries mandate or incentivize digital platforms to deploy machine learning to remove disfavored political, social, and religious speech.” For example, the Russian Federal Service for Supervision of Communications, Information Technology and Mass Media (‘Roskomnadzor’) openly uses ML tools to filter online content, identifying ‘unwanted’ content with a reportedly high degree of accuracy \citep{Gaufman2021Cybercrime}. CoPE's improved efficiency and accuracy could make such systems more effective.

To mitigate this risk, we suggest that policy interpretation models consider using a variation of the OpenRAIL-AMS license when released publicly \citep{Contractor2022RAIL}. Specifically, licenses could include mandatory use-based restrictions that explicitly prohibit mass surveillance and protect user privacy, drawing definitions from UN Special Rapporteur guidance. The license requires downstream users to maintain these constraints and holds users accountable for model outputs, while granting the licensor powers to restrict or update artifacts if violations are detected. We recognize the enforcement challenges of these mechanisms \citep{downing2023licensing}, but believe, on balance, explicit legal restrictions can create more accountability in places where rule of law is respected. 

\subsection{Centralization vs. Pluralism}
A related concern is that improving moderation efficiency might entrench the values of large platforms. However, the steerability property of CoPE benefits smaller and decentralized platforms—subreddits, Bluesky, Discord communities—which lack resources for sophisticated moderation at scale. CoPE's policy-agnostic design enables these actors to define their own standards rather than adopting defaults from larger systems. Whether this produces greater diversity in practice depends on adoption patterns we cannot predict \citep{noels2025largelanguagemodelstalk}.

\subsection{Language Limitations}
CoPE is trained and tested exclusively on English-language content. While the base Gemma 2 model supports other languages, we have not validated performance in non-English contexts. This choice simplified our initial validation but perpetuates English-language dominance in content moderation tooling. Organizations enforcing policies in other languages face a choice: use an unvalidated system or continue with less effective tools \citep{peppin2025multilingualdivideimpactglobal}. We plan to expand training and testing to other languages, but currently this remains a significant limitation. 

\subsection{Bias and Fairness}
Content moderation models can exhibit multiple types of bias, including oversensitivity to certain words, differential performance across demographic groups, and problematic topic associations \citep{10.1145/3726986.3727014, doi:10.1073/pnas.2416228122}. \citet{wang2025probingassociationbiasesllm} suggest that embedded biases may be attached to topic associations, rather than the prevalence of certain language. CoPE inherits these risks from its training process and base model. We have not conducted systematic fairness audits across demographic groups or cultural contexts. The contradictory training methodology may reduce some biases by preventing the model from learning fixed cultural heuristics, but this remains an empirical question requiring further investigation.

\subsection{Generative Risks}
CoPE's fine-tuning for classification limits certain risks associated with general-purpose LLMs. It cannot generate harmful advice or persuasive misinformation in the way conversational models can. However, it introduces different risks through its potential to scale enforcement of harmful policies. Unlike general-purpose models where harm often comes from individual interactions, moderation systems affect large populations through systematic enforcement decisions.

The open-weight release of CoPE-A reflects our position that progress in content moderation benefits from collaborative development and public scrutiny. By releasing the model, methodology, and evaluation framework, we enable independent assessment of these risks and community-driven improvements.

\section{Future Work}

\subsection{Explanations}
While the current CoPE model reliably generates classification labels aligned with moderation policies, it does not yet provide explicit rationales explaining these policy decisions. Future work should therefore prioritize enhancing CoPE’s explainability, allowing moderators,  policy-makers, and people subject to moderation to better understand how automated moderation decisions are made and facilitating more informed human responses and governance. We believe we should be able to train for this behavior by augmenting our existing dataset with case-by-case explanations and will be investigating approaches to do so scalably over the next year. Such improvements would not only enhance usability for trust and safety teams but also promote greater transparency and trust in automated moderation systems.   

\subsection{Modalities and Languages}
Content labeling at scale requires the ability to assess multimodal content, including text, images, audio, and video, across diverse languages, cultures, and evolving harm taxonomies. Future iterations of CoPE should work to integrate multimodal processing capabilities, thus expanding the model’s applicability to a broader range of online content generated by human users and machines. Additionally, extending multilingual and multicultural coverage beyond English to accommodate global needs will constitute another critical step in achieving generalizability. Moreover, continuously training the model on updated policies and content samples that reflect emerging harms will ensure CoPE remains adaptable to the rapidly evolving content ecosystem.

\subsection{Feedback Loops}
Content moderation systems need to be continuously improved and inspected, since they deal with the delicate boundaries between freedom of speech versus a responsible online space. Therefore, future iterations of CoPE should include the development of a systematic feedback loop incorporating real-time human moderator feedback, user appeals data, and moderation decision collection. Allowing moderators and end users to flag uncertain or incorrect judgments will not only enhance model accuracy but also ensure CoPE remains aligned with pluralistic moral principles over time, effectively adapting to societal norms and regulatory standards.

\section{Conclusion}

This paper demonstrates that CoPE– a 9-billion parameter model fine-tuned on contradictory policy examples– can match or exceed the content labeling accuracy of frontier models across multiple harm categories. CoPE achieves this with a ~100x reduction in model size compared to GPT-4, with corresponding improvements in deployment cost and latency. We believe this combination of steerability, accuracy, and efficiency represents a new era for content labeling technology. Organizations can use CoPE to update policies without model retraining, enabling unprecedented moderation agility. Different communities can enforce different standards using the same infrastructure, supporting true policy pluralism. And policy-explicit labeling can make moderation criteria transparent, ultimately enabling greater accountability in online governance.

\begin{acks}
We would like to thank Alex Rosenblatt, Eyal Zilberman, Jiaao Chen, Sophie Duba, Vibhor Kumar, and Welton Wang for their contributions to this project. We are grateful to Stanford for its support of this project and to OpenAI and SafetyKit for their in-kind support. 
\end{acks}



\begin{thebibliography}{43}


\ifx \showCODEN    \undefined \def \showCODEN     #1{\unskip}     \fi
\ifx \showISBNx    \undefined \def \showISBNx     #1{\unskip}     \fi
\ifx \showISBNxiii \undefined \def \showISBNxiii  #1{\unskip}     \fi
\ifx \showISSN     \undefined \def \showISSN      #1{\unskip}     \fi
\ifx \showLCCN     \undefined \def \showLCCN      #1{\unskip}     \fi
\ifx \shownote     \undefined \def \shownote      #1{#1}          \fi
\ifx \showarticletitle \undefined \def \showarticletitle #1{#1}   \fi
\ifx \showURL      \undefined \def \showURL       {\relax}        \fi
\providecommand\bibfield[2]{#2}
\providecommand\bibinfo[2]{#2}
\providecommand\natexlab[1]{#1}
\providecommand\showeprint[2][]{arXiv:#2}

\bibitem[Bai et~al\mbox{.}(2025)]%
        {doi:10.1073/pnas.2416228122}
\bibfield{author}{\bibinfo{person}{Xuechunzi Bai}, \bibinfo{person}{Angelina Wang}, \bibinfo{person}{Ilia Sucholutsky}, {and} \bibinfo{person}{Thomas~L. Griffiths}.} \bibinfo{year}{2025}\natexlab{}.
\newblock \showarticletitle{Explicitly unbiased large language models still form biased associations}.
\newblock \bibinfo{journal}{\emph{Proceedings of the National Academy of Sciences}} \bibinfo{volume}{122}, \bibinfo{number}{8} (\bibinfo{year}{2025}), \bibinfo{pages}{e2416228122}.
\newblock
\showeprint{https://www.pnas.org/doi/pdf/10.1073/pnas.2416228122}
\href{https://doi.org/10.1073/pnas.2416228122}{doi:\nolinkurl{10.1073/pnas.2416228122}}


\bibitem[Caplan(2018)]%
        {caplan2018contentOrContextModeration}
\bibfield{author}{\bibinfo{person}{Robyn Caplan}.} \bibinfo{year}{2018}\natexlab{}.
\newblock \bibinfo{booktitle}{\emph{Content or Context Moderation?}}
\newblock \bibinfo{type}{{T}echnical {R}eport}. \bibinfo{institution}{Data \& Society}.
\newblock
\urldef\tempurl%
\url{https://datasociety.net/library/content-or-context-moderation/}
\showURL{%
\tempurl}
\newblock
\shownote{Technical report}.


\bibitem[Chi et~al\mbox{.}(2024)]%
        {chi2024llamaGuard3Vision}
\bibfield{author}{\bibinfo{person}{Jinhao Chi}, \bibinfo{person}{Upendra Karn}, \bibinfo{person}{Hanlin Zhan}, \bibinfo{person}{Eric Smith}, \bibinfo{person}{Jared Rando}, \bibinfo{person}{Yuzhe Zhang}, \bibinfo{person}{Kamil Plawiak}, \bibinfo{person}{Zachary~D. Coudert}, \bibinfo{person}{Kartik Upasani}, {and} \bibinfo{person}{Manish Pasupuleti}.} \bibinfo{year}{2024}\natexlab{}.
\newblock \bibinfo{title}{Llama Guard 3 Vision: Safeguarding Human-{AI} Image Understanding Conversations}.
\newblock \bibinfo{howpublished}{arXiv}.
\newblock
\urldef\tempurl%
\url{https://arxiv.org/abs/2411.10414}
\showURL{%
\tempurl}
\newblock
\shownote{Date in source list: 2024-11-15}.


\bibitem[Contractor et~al\mbox{.}(2022)]%
        {Contractor2022RAIL}
\bibfield{author}{\bibinfo{person}{Danish Contractor}, \bibinfo{person}{Carlos~Munoz Ferrandis}, \bibinfo{person}{Jenny Lee}, {and} \bibinfo{person}{Daniel McDuff}.} \bibinfo{year}{2022}\natexlab{}.
\newblock \bibinfo{title}{From {RAIL} to {Open RAIL}: {T}opologies of {RAIL} Licenses}.
\newblock \bibinfo{howpublished}{Responsible AI Licenses (RAIL)}.
\newblock
\urldef\tempurl%
\url{https://www.licenses.ai/blog/2022/8/18/naming-convention-of-responsible-ai-licenses}
\showURL{%
\tempurl}
\newblock
\shownote{Accessed: 2025-12-18}.


\bibitem[Downing(2023)]%
        {downing2023licensing}
\bibfield{author}{\bibinfo{person}{Kate Downing}.} \bibinfo{year}{2023}\natexlab{}.
\newblock \bibinfo{title}{AI Licensing Can’t Balance “Open” with “Responsible”}.
\newblock
\urldef\tempurl%
\url{https://katedowninglaw.com/2023/07/13/ai-licensing-cant-balance-open-with-responsible/}
\showURL{%
\tempurl}


\bibitem[Earl et~al\mbox{.}(2022)]%
        {doi:10.1126/sciadv.abl8198}
\bibfield{author}{\bibinfo{person}{Jennifer Earl}, \bibinfo{person}{Thomas~V. Maher}, {and} \bibinfo{person}{Jennifer Pan}.} \bibinfo{year}{2022}\natexlab{}.
\newblock \showarticletitle{The digital repression of social movements, protest, and activism: A synthetic review}.
\newblock \bibinfo{journal}{\emph{Science Advances}} \bibinfo{volume}{8}, \bibinfo{number}{10} (\bibinfo{year}{2022}), \bibinfo{pages}{eabl8198}.
\newblock
\showeprint{https://www.science.org/doi/pdf/10.1126/sciadv.abl8198}
\href{https://doi.org/10.1126/sciadv.abl8198}{doi:\nolinkurl{10.1126/sciadv.abl8198}}


\bibitem[Engstrom and Feamster(2017)]%
        {engstrom2017limitsFiltering}
\bibfield{author}{\bibinfo{person}{Evan Engstrom} {and} \bibinfo{person}{Nick Feamster}.} \bibinfo{year}{2017}\natexlab{}.
\newblock \bibinfo{booktitle}{\emph{The Limits of Filtering: A Look at the Functionality \& Shortcomings of Content Detection Tools}}.
\newblock \bibinfo{type}{{T}echnical {R}eport}. \bibinfo{institution}{Engine}.
\newblock
\urldef\tempurl%
\url{https://www.engine.is/the-limits-of-filtering}
\showURL{%
\tempurl}
\newblock
\shownote{Technical report}.


\bibitem[Funk et~al\mbox{.}(2023)]%
        {Funk2023Repressive}
\bibfield{author}{\bibinfo{person}{Allie Funk}, \bibinfo{person}{Adrian Shahbaz}, {and} \bibinfo{person}{Kian Vesteinsson}.} \bibinfo{year}{2023}\natexlab{}.
\newblock \showarticletitle{The Repressive Power of Artificial Intelligence}.
\newblock In \bibinfo{booktitle}{\emph{Freedom on the Net 2023: The Repressive Power of Artificial Intelligence}}, \bibfield{editor}{\bibinfo{person}{Allie Funk}, \bibinfo{person}{Adrian Shahbaz}, \bibinfo{person}{Kian Vesteinsson}, \bibinfo{person}{Jennifer Brody}, \bibinfo{person}{Grant Baker}, \bibinfo{person}{Cathryn Grothe}, \bibinfo{person}{Matthew Barak}, \bibinfo{person}{Maddie Masinsin}, \bibinfo{person}{Rucha Modi}, {and} \bibinfo{person}{Elizabeth Sutterlin}} (Eds.). \bibinfo{publisher}{Freedom House}.
\newblock
\urldef\tempurl%
\url{https://freedomhouse.org/report/freedom-net/2023/repressive-power-artificial-intelligence}
\showURL{%
\tempurl}
\newblock
\shownote{Available at freedomonthenet.org}.


\bibitem[Gaufman(2021)]%
        {Gaufman2021Cybercrime}
\bibfield{author}{\bibinfo{person}{Elizaveta Gaufman}.} \bibinfo{year}{2021}\natexlab{}.
\newblock \showarticletitle{Cybercrime and Punishment: Security, Information War, and the Future of Runet}.
\newblock In \bibinfo{booktitle}{\emph{The Palgrave Handbook of Digital Russia Studies}}, \bibfield{editor}{\bibinfo{person}{Daria Gritsenko}, \bibinfo{person}{Mari{\"e}lle Wijermars}, {and} \bibinfo{person}{Mikhail Kopotev}} (Eds.). \bibinfo{publisher}{Springer International Publishing}, \bibinfo{address}{Cham}, \bibinfo{pages}{115--134}.
\newblock
\showISBNx{978-3-030-42855-6}
\href{https://doi.org/10.1007/978-3-030-42855-6_7}{doi:\nolinkurl{10.1007/978-3-030-42855-6_7}}


\bibitem[Gomez et~al\mbox{.}(2024)]%
        {gomez2024algorithmicArbitrariness}
\bibfield{author}{\bibinfo{person}{Jo{\~a}o~F. Gomez}, \bibinfo{person}{C. Vieira~Machado}, \bibinfo{person}{L. Monteiro~Paes}, {and} \bibinfo{person}{Fl{\'a}vio~P. Calmon}.} \bibinfo{year}{2024}\natexlab{}.
\newblock \bibinfo{title}{Algorithmic Arbitrariness in Content Moderation}.
\newblock \bibinfo{howpublished}{arXiv}.
\newblock
\urldef\tempurl%
\url{https://arxiv.org/abs/2402.16979}
\showURL{%
\tempurl}


\bibitem[Han et~al\mbox{.}(2024)]%
        {10.5555/3737916.3738177}
\bibfield{author}{\bibinfo{person}{Seungju Han}, \bibinfo{person}{Kavel Rao}, \bibinfo{person}{Allyson Ettinger}, \bibinfo{person}{Liwei Jiang}, \bibinfo{person}{Bill~Yuchen Lin}, \bibinfo{person}{Nathan Lambert}, \bibinfo{person}{Yejin Choi}, {and} \bibinfo{person}{Nouha Dziri}.} \bibinfo{year}{2024}\natexlab{}.
\newblock \showarticletitle{WILDGUARD: open one-stop moderation tools for safety risks, jailbreaks, and refusals of LLMs}. In \bibinfo{booktitle}{\emph{Proceedings of the 38th International Conference on Neural Information Processing Systems}} (Vancouver, BC, Canada) \emph{(\bibinfo{series}{NIPS '24})}. \bibinfo{publisher}{Curran Associates Inc.}, \bibinfo{address}{Red Hook, NY, USA}, Article \bibinfo{articleno}{261}, \bibinfo{numpages}{39}~pages.
\newblock
\showISBNx{9798331314385}


\bibitem[Inan et~al\mbox{.}(2023)]%
        {inan2023llamaguardllmbasedinputoutput}
\bibfield{author}{\bibinfo{person}{Hakan Inan}, \bibinfo{person}{Kartikeya Upasani}, \bibinfo{person}{Jianfeng Chi}, \bibinfo{person}{Rashi Rungta}, \bibinfo{person}{Krithika Iyer}, \bibinfo{person}{Yuning Mao}, \bibinfo{person}{Michael Tontchev}, \bibinfo{person}{Qing Hu}, \bibinfo{person}{Brian Fuller}, \bibinfo{person}{Davide Testuggine}, {and} \bibinfo{person}{Madian Khabsa}.} \bibinfo{year}{2023}\natexlab{}.
\newblock \bibinfo{title}{Llama Guard: LLM-based Input-Output Safeguard for Human-AI Conversations}.
\newblock
\showeprint[arxiv]{2312.06674}~[cs.CL]
\urldef\tempurl%
\url{https://arxiv.org/abs/2312.06674}
\showURL{%
\tempurl}


\bibitem[Jhaver et~al\mbox{.}(2019)]%
        {jhaver2019automoderator}
\bibfield{author}{\bibinfo{person}{Shagun Jhaver}, \bibinfo{person}{Israel Birman}, \bibinfo{person}{Eric Gilbert}, {and} \bibinfo{person}{Amy Bruckman}.} \bibinfo{year}{2019}\natexlab{}.
\newblock \showarticletitle{Human-Machine Collaboration for Content Regulation: The Case of Reddit Automoderator}.
\newblock \bibinfo{journal}{\emph{ACM Transactions on Computer-Human Interaction}} \bibinfo{volume}{26}, \bibinfo{number}{5} (\bibinfo{year}{2019}), \bibinfo{pages}{1--35}.
\newblock
\href{https://doi.org/10.1145/3338243}{doi:\nolinkurl{10.1145/3338243}}


\bibitem[Kholkar and Ahuja(2025)]%
        {kholkar2025policyaspromptturningaigovernance}
\bibfield{author}{\bibinfo{person}{Gauri Kholkar} {and} \bibinfo{person}{Ratinder Ahuja}.} \bibinfo{year}{2025}\natexlab{}.
\newblock \bibinfo{title}{Policy-as-Prompt: Turning AI Governance Rules into Guardrails for AI Agents}.
\newblock
\showeprint[arxiv]{2509.23994}~[cs.CL]
\urldef\tempurl%
\url{https://arxiv.org/abs/2509.23994}
\showURL{%
\tempurl}


\bibitem[Kumar et~al\mbox{.}(2025)]%
        {kumar2025noFreeLunchGuardrails}
\bibfield{author}{\bibinfo{person}{Deepak Kumar}, \bibinfo{person}{N.~A. Birur}, \bibinfo{person}{T. Baswa}, \bibinfo{person}{Sahil Agarwal}, {and} \bibinfo{person}{Prashanth Harshangi}.} \bibinfo{year}{2025}\natexlab{}.
\newblock \bibinfo{title}{No Free Lunch with Guardrails}.
\newblock \bibinfo{howpublished}{arXiv}.
\newblock
\urldef\tempurl%
\url{https://arxiv.org/abs/2504.00441}
\showURL{%
\tempurl}


\bibitem[Li et~al\mbox{.}(2024)]%
        {li2024safetyAnalyst}
\bibfield{author}{\bibinfo{person}{Jasmine Li}, \bibinfo{person}{Valentina Pyatkin}, \bibinfo{person}{Max Kleiman-Weiner}, \bibinfo{person}{Liwei Jiang}, \bibinfo{person}{Nouha Dziri}, \bibinfo{person}{Anne G.~E. Collins}, \bibinfo{person}{Jana~S. Borg}, \bibinfo{person}{Maarten Sap}, \bibinfo{person}{Yejin Choi}, {and} \bibinfo{person}{Sergey Levine}.} \bibinfo{year}{2024}\natexlab{}.
\newblock \bibinfo{title}{SafetyAnalyst: Interpretable, Transparent, and Steerable safety moderation for {AI} behavior}.
\newblock \bibinfo{howpublished}{arXiv}.
\newblock
\urldef\tempurl%
\url{https://arxiv.org/abs/2410.16665}
\showURL{%
\tempurl}
\newblock
\shownote{Project page: https://jl3676.github.io/SafetyAnalyst/}.


\bibitem[Liu et~al\mbox{.}(2025)]%
        {liu2025guardreasoner}
\bibfield{author}{\bibinfo{person}{Yizhou Liu}, \bibinfo{person}{Hongcheng Gao}, \bibinfo{person}{Siyi Zhai}, \bibinfo{person}{Yuhang He}, \bibinfo{person}{Jiacheng Xia}, \bibinfo{person}{Zhen Hu}, \bibinfo{person}{Yuxuan Chen}, \bibinfo{person}{Xin Yang}, \bibinfo{person}{Jie Zhang}, \bibinfo{person}{Shiliang~Z. Li}, \bibinfo{person}{Hui Xiong}, {and} \bibinfo{person}{Bryan Hooi}.} \bibinfo{year}{2025}\natexlab{}.
\newblock \bibinfo{title}{GuardReasoner: Towards Reasoning-based {LLM} safeguards}.
\newblock \bibinfo{howpublished}{arXiv}.
\newblock
\urldef\tempurl%
\url{https://arxiv.org/abs/2501.18492}
\showURL{%
\tempurl}


\bibitem[Lu et~al\mbox{.}(2025)]%
        {lu-etal-2025-llm}
\bibfield{author}{\bibinfo{person}{Junyu Lu}, \bibinfo{person}{Kai Ma}, \bibinfo{person}{Kaichun Wang}, \bibinfo{person}{Kelaiti Xiao}, \bibinfo{person}{Roy Ka-Wei Lee}, \bibinfo{person}{Bo Xu}, \bibinfo{person}{Liang Yang}, {and} \bibinfo{person}{Hongfei Lin}.} \bibinfo{year}{2025}\natexlab{}.
\newblock \showarticletitle{Is {LLM} an Overconfident Judge? Unveiling the Capabilities of {LLM}s in Detecting Offensive Language with Annotation Disagreement}. In \bibinfo{booktitle}{\emph{Findings of the Association for Computational Linguistics: ACL 2025}}, \bibfield{editor}{\bibinfo{person}{Wanxiang Che}, \bibinfo{person}{Joyce Nabende}, \bibinfo{person}{Ekaterina Shutova}, {and} \bibinfo{person}{Mohammad~Taher Pilehvar}} (Eds.). \bibinfo{publisher}{Association for Computational Linguistics}, \bibinfo{address}{Vienna, Austria}, \bibinfo{pages}{5609--5626}.
\newblock
\showISBNx{979-8-89176-256-5}
\href{https://doi.org/10.18653/v1/2025.findings-acl.293}{doi:\nolinkurl{10.18653/v1/2025.findings-acl.293}}


\bibitem[Ma et~al\mbox{.}(2024)]%
        {ma2024adaptinglargelanguagemodels}
\bibfield{author}{\bibinfo{person}{Huan Ma}, \bibinfo{person}{Changqing Zhang}, \bibinfo{person}{Huazhu Fu}, \bibinfo{person}{Peilin Zhao}, {and} \bibinfo{person}{Bingzhe Wu}.} \bibinfo{year}{2024}\natexlab{}.
\newblock \bibinfo{title}{Adapting Large Language Models for Content Moderation: Pitfalls in Data Engineering and Supervised Fine-tuning}.
\newblock
\showeprint[arxiv]{2310.03400}~[cs.LG]
\urldef\tempurl%
\url{https://arxiv.org/abs/2310.03400}
\showURL{%
\tempurl}


\bibitem[Markov et~al\mbox{.}(2023)]%
        {10.1609/aaai.v37i12.26752}
\bibfield{author}{\bibinfo{person}{Todor Markov}, \bibinfo{person}{Chong Zhang}, \bibinfo{person}{Sandhini Agarwal}, \bibinfo{person}{Florentine~Eloundou Nekoul}, \bibinfo{person}{Theodore Lee}, \bibinfo{person}{Steven Adler}, \bibinfo{person}{Angela Jiang}, {and} \bibinfo{person}{Lilian Weng}.} \bibinfo{year}{2023}\natexlab{}.
\newblock \showarticletitle{A holistic approach to undesired content detection in the real world}. In \bibinfo{booktitle}{\emph{Proceedings of the Thirty-Seventh AAAI Conference on Artificial Intelligence and Thirty-Fifth Conference on Innovative Applications of Artificial Intelligence and Thirteenth Symposium on Educational Advances in Artificial Intelligence}} \emph{(\bibinfo{series}{AAAI'23/IAAI'23/EAAI'23})}. \bibinfo{publisher}{AAAI Press}, Article \bibinfo{articleno}{1683}, \bibinfo{numpages}{10}~pages.
\newblock
\showISBNx{978-1-57735-880-0}
\href{https://doi.org/10.1609/aaai.v37i12.26752}{doi:\nolinkurl{10.1609/aaai.v37i12.26752}}


\bibitem[Masnick(2019)]%
        {masnick2019impossibility}
\bibfield{author}{\bibinfo{person}{Mike Masnick}.} \bibinfo{year}{2019}\natexlab{}.
\newblock \bibinfo{booktitle}{\emph{Masnick’s Impossibility Theorem: Content Moderation at Scale Is Impossible To Do Well}}.
\newblock Techdirt.
\newblock
\urldef\tempurl%
\url{https://www.techdirt.com/2019/11/20/masnicks-impossibility-theorem-content-moderation-scale-is-impossible-to-do-well/}
\showURL{%
\tempurl}
\newblock
\shownote{Accessed: 2025-07-14}.


\bibitem[{Mistral AI}(2025)]%
        {mistral2025guardrailingDocs}
\bibfield{author}{\bibinfo{person}{{Mistral AI}}.} \bibinfo{year}{2025}\natexlab{}.
\newblock \bibinfo{booktitle}{\emph{Moderation \& Guardrailing | Mistral Docs}}.
\newblock
\urldef\tempurl%
\url{https://docs.mistral.ai/capabilities/guardrailing/}
\showURL{%
\tempurl}


\bibitem[Mollas et~al\mbox{.}(2022)]%
        {mollas2022ethos}
\bibfield{author}{\bibinfo{person}{Ioannis Mollas}, \bibinfo{person}{Zampia Chrysopoulou}, \bibinfo{person}{Stamatis Karlos}, {et~al\mbox{.}}} \bibinfo{year}{2022}\natexlab{}.
\newblock \showarticletitle{{ETHOS}: a multi-label hate speech detection dataset}.
\newblock \bibinfo{journal}{\emph{Complex \& Intelligent Systems}}  \bibinfo{volume}{8} (\bibinfo{year}{2022}), \bibinfo{pages}{4663--4678}.
\newblock
\href{https://doi.org/10.1007/s40747-021-00608-2}{doi:\nolinkurl{10.1007/s40747-021-00608-2}}


\bibitem[Noels et~al\mbox{.}(2025)]%
        {noels2025largelanguagemodelstalk}
\bibfield{author}{\bibinfo{person}{Sander Noels}, \bibinfo{person}{Guillaume Bied}, \bibinfo{person}{Maarten Buyl}, \bibinfo{person}{Alexander Rogiers}, \bibinfo{person}{Yousra Fettach}, \bibinfo{person}{Jefrey Lijffijt}, {and} \bibinfo{person}{Tijl~De Bie}.} \bibinfo{year}{2025}\natexlab{}.
\newblock \bibinfo{title}{What Large Language Models Do Not Talk About: An Empirical Study of Moderation and Censorship Practices}.
\newblock
\showeprint[arxiv]{2504.03803}~[cs.CL]
\urldef\tempurl%
\url{https://arxiv.org/abs/2504.03803}
\showURL{%
\tempurl}


\bibitem[{OpenAI}(2025a)]%
        {openai2025gptOssSafeguard}
\bibfield{author}{\bibinfo{person}{{OpenAI}}.} \bibinfo{year}{2025}\natexlab{a}.
\newblock \bibinfo{booktitle}{\emph{Introducing gpt-oss-safeguard}}.
\newblock
\urldef\tempurl%
\url{https://openai.com/index/introducing-gpt-oss-safeguard/}
\showURL{%
\tempurl}


\bibitem[{OpenAI}(2025b)]%
        {openai2025moderationDocs}
\bibfield{author}{\bibinfo{person}{{OpenAI}}.} \bibinfo{year}{2025}\natexlab{b}.
\newblock \bibinfo{booktitle}{\emph{Moderation | OpenAI Platform Docs}}.
\newblock
\urldef\tempurl%
\url{https://platform.openai.com/docs/guides/moderation}
\showURL{%
\tempurl}


\bibitem[Padhi et~al\mbox{.}(2025)]%
        {padhi-etal-2025-granite}
\bibfield{author}{\bibinfo{person}{Inkit Padhi}, \bibinfo{person}{Manish Nagireddy}, \bibinfo{person}{Giandomenico Cornacchia}, \bibinfo{person}{Subhajit Chaudhury}, \bibinfo{person}{Tejaswini Pedapati}, \bibinfo{person}{Pierre Dognin}, \bibinfo{person}{Keerthiram Murugesan}, \bibinfo{person}{Erik Miehling}, \bibinfo{person}{Mart{\'i}n Santill{\'a}n~Cooper}, \bibinfo{person}{Kieran Fraser}, \bibinfo{person}{Giulio Zizzo}, \bibinfo{person}{Muhammad~Zaid Hameed}, \bibinfo{person}{Mark Purcell}, \bibinfo{person}{Michael Desmond}, \bibinfo{person}{Qian Pan}, \bibinfo{person}{Inge Vejsbjerg}, \bibinfo{person}{Elizabeth~M. Daly}, \bibinfo{person}{Michael Hind}, \bibinfo{person}{Werner Geyer}, \bibinfo{person}{Ambrish Rawat}, \bibinfo{person}{Kush~R. Varshney}, {and} \bibinfo{person}{Prasanna Sattigeri}.} \bibinfo{year}{2025}\natexlab{}.
\newblock \showarticletitle{Granite Guardian: Comprehensive {LLM} Safeguarding}. In \bibinfo{booktitle}{\emph{Proceedings of the 2025 Conference of the Nations of the Americas Chapter of the Association for Computational Linguistics: Human Language Technologies (Volume 3: Industry Track)}}, \bibfield{editor}{\bibinfo{person}{Weizhu Chen}, \bibinfo{person}{Yi~Yang}, \bibinfo{person}{Mohammad Kachuee}, {and} \bibinfo{person}{Xue-Yong Fu}} (Eds.). \bibinfo{publisher}{Association for Computational Linguistics}, \bibinfo{address}{Albuquerque, New Mexico}, \bibinfo{pages}{607--615}.
\newblock
\showISBNx{979-8-89176-194-0}
\href{https://doi.org/10.18653/v1/2025.naacl-industry.49}{doi:\nolinkurl{10.18653/v1/2025.naacl-industry.49}}


\bibitem[Palla et~al\mbox{.}(2025)]%
        {palla2025policyAsPrompt}
\bibfield{author}{\bibinfo{person}{Konstantinos Palla}, \bibinfo{person}{Jos{\'e} L.~R. Garc{\'\i}a}, \bibinfo{person}{Claudia Hauff}, \bibinfo{person}{Federica Fabbri}, \bibinfo{person}{Helge Lindstr{\"o}m}, \bibinfo{person}{D.~R. Taber}, \bibinfo{person}{Andreas Damianou}, {and} \bibinfo{person}{Mounia Lalmas}.} \bibinfo{year}{2025}\natexlab{}.
\newblock \showarticletitle{Policy-as-Prompt: Rethinking content moderation in the age of large language models}. In \bibinfo{booktitle}{\emph{Proceedings of the 2025 ACM Conference on Fairness, Accountability, and Transparency (FAccT '25)}}. \bibinfo{publisher}{Association for Computing Machinery}, \bibinfo{address}{New York, NY, USA}, \bibinfo{pages}{840--854}.
\newblock
\href{https://doi.org/10.1145/3715275.3732054}{doi:\nolinkurl{10.1145/3715275.3732054}}


\bibitem[Peppin et~al\mbox{.}(2025)]%
        {peppin2025multilingualdivideimpactglobal}
\bibfield{author}{\bibinfo{person}{Aidan Peppin}, \bibinfo{person}{Julia Kreutzer}, \bibinfo{person}{Alice~Schoenauer Sebag}, \bibinfo{person}{Kelly Marchisio}, \bibinfo{person}{Beyza Ermis}, \bibinfo{person}{John Dang}, \bibinfo{person}{Samuel Cahyawijaya}, \bibinfo{person}{Shivalika Singh}, \bibinfo{person}{Seraphina Goldfarb-Tarrant}, \bibinfo{person}{Viraat Aryabumi}, \bibinfo{person}{Aakanksha}, \bibinfo{person}{Wei-Yin Ko}, \bibinfo{person}{Ahmet Üstün}, \bibinfo{person}{Matthias Gallé}, \bibinfo{person}{Marzieh Fadaee}, {and} \bibinfo{person}{Sara Hooker}.} \bibinfo{year}{2025}\natexlab{}.
\newblock \bibinfo{title}{The Multilingual Divide and Its Impact on Global AI Safety}.
\newblock
\showeprint[arxiv]{2505.21344}~[cs.AI]
\urldef\tempurl%
\url{https://arxiv.org/abs/2505.21344}
\showURL{%
\tempurl}


\bibitem[Ray(2024)]%
        {ray2024policyAsCode}
\bibfield{author}{\bibinfo{person}{Justin Ray}.} \bibinfo{year}{2024}\natexlab{}.
\newblock \bibinfo{booktitle}{\emph{Policy as Code: Improving Cloud Native Security}}.
\newblock \bibinfo{publisher}{O’Reilly Media}.
\newblock
\showISBNx{9781098139186}


\bibitem[Reda(2017)]%
        {reda2017whenFiltersFail}
\bibfield{author}{\bibinfo{person}{Felix Reda}.} \bibinfo{year}{2017}\natexlab{}.
\newblock \bibinfo{booktitle}{\emph{When filters fail: These cases show we can’t trust algorithms to clean up the internet}}.
\newblock
\urldef\tempurl%
\url{https://felixreda.eu/2017/09/when-filters-fail/}
\showURL{%
\tempurl}


\bibitem[Roberts(2019)]%
        {roberts2019behindTheScreen}
\bibfield{author}{\bibinfo{person}{Sarah~T. Roberts}.} \bibinfo{year}{2019}\natexlab{}.
\newblock \bibinfo{booktitle}{\emph{Behind the Screen: Content Moderation in the Shadows of Social Media}}.
\newblock \bibinfo{publisher}{Yale University Press}.
\newblock
\urldef\tempurl%
\url{https://www.jstor.org/stable/j.ctvhrcz0v}
\showURL{%
\tempurl}
\newblock
\shownote{JSTOR access link}.


\bibitem[Shi et~al\mbox{.}(2025)]%
        {shi2025progentprogrammableprivilegecontrol}
\bibfield{author}{\bibinfo{person}{Tianneng Shi}, \bibinfo{person}{Jingxuan He}, \bibinfo{person}{Zhun Wang}, \bibinfo{person}{Hongwei Li}, \bibinfo{person}{Linyu Wu}, \bibinfo{person}{Wenbo Guo}, {and} \bibinfo{person}{Dawn Song}.} \bibinfo{year}{2025}\natexlab{}.
\newblock \bibinfo{title}{Progent: Programmable Privilege Control for LLM Agents}.
\newblock
\showeprint[arxiv]{2504.11703}~[cs.CR]
\urldef\tempurl%
\url{https://arxiv.org/abs/2504.11703}
\showURL{%
\tempurl}


\bibitem[Son et~al\mbox{.}(2023)]%
        {son2023contentModerationWild}
\bibfield{author}{\bibinfo{person}{Donghyun Son}, \bibinfo{person}{Byungkeun Lew}, \bibinfo{person}{Kyunghwan Choi}, \bibinfo{person}{Yoon Baek}, \bibinfo{person}{Seonghyeon Choi}, \bibinfo{person}{Byung Shin}, \bibinfo{person}{Sangjun Ha}, {and} \bibinfo{person}{Byoungju Chang}.} \bibinfo{year}{2023}\natexlab{}.
\newblock \bibinfo{title}{Reliable Decision from Multiple Subtasks through Threshold Optimization: Content Moderation in the Wild}.
\newblock \bibinfo{howpublished}{arXiv}.
\newblock
\urldef\tempurl%
\url{https://arxiv.org/pdf/2208.07522}
\showURL{%
\tempurl}
\newblock
\shownote{arXiv preprint; URL provided points to PDF}.


\bibitem[Team et~al\mbox{.}(2024)]%
        {gemmateam2024gemma2improvingopen}
\bibfield{author}{\bibinfo{person}{Gemma Team}, \bibinfo{person}{Morgane Riviere}, \bibinfo{person}{Shreya Pathak}, \bibinfo{person}{Pier~Giuseppe Sessa}, \bibinfo{person}{Cassidy Hardin}, \bibinfo{person}{Surya Bhupatiraju}, \bibinfo{person}{Léonard Hussenot}, \bibinfo{person}{Thomas Mesnard}, \bibinfo{person}{Bobak Shahriari}, \bibinfo{person}{Alexandre Ramé}, \bibinfo{person}{Johan Ferret}, \bibinfo{person}{Peter Liu}, \bibinfo{person}{Pouya Tafti}, \bibinfo{person}{Abe Friesen}, \bibinfo{person}{Michelle Casbon}, \bibinfo{person}{Sabela Ramos}, \bibinfo{person}{Ravin Kumar}, \bibinfo{person}{Charline~Le Lan}, \bibinfo{person}{Sammy Jerome}, \bibinfo{person}{Anton Tsitsulin}, \bibinfo{person}{Nino Vieillard}, \bibinfo{person}{Piotr Stanczyk}, \bibinfo{person}{Sertan Girgin}, \bibinfo{person}{Nikola Momchev}, \bibinfo{person}{Matt Hoffman}, \bibinfo{person}{Shantanu Thakoor}, \bibinfo{person}{Jean-Bastien Grill}, \bibinfo{person}{Behnam Neyshabur}, \bibinfo{person}{Olivier Bachem}, \bibinfo{person}{Alanna
  Walton}, \bibinfo{person}{Aliaksei Severyn}, \bibinfo{person}{Alicia Parrish}, \bibinfo{person}{Aliya Ahmad}, \bibinfo{person}{Allen Hutchison}, \bibinfo{person}{Alvin Abdagic}, \bibinfo{person}{Amanda Carl}, \bibinfo{person}{Amy Shen}, \bibinfo{person}{Andy Brock}, \bibinfo{person}{Andy Coenen}, \bibinfo{person}{Anthony Laforge}, \bibinfo{person}{Antonia Paterson}, \bibinfo{person}{Ben Bastian}, \bibinfo{person}{Bilal Piot}, \bibinfo{person}{Bo Wu}, \bibinfo{person}{Brandon Royal}, \bibinfo{person}{Charlie Chen}, \bibinfo{person}{Chintu Kumar}, \bibinfo{person}{Chris Perry}, \bibinfo{person}{Chris Welty}, \bibinfo{person}{Christopher~A. Choquette-Choo}, \bibinfo{person}{Danila Sinopalnikov}, \bibinfo{person}{David Weinberger}, \bibinfo{person}{Dimple Vijaykumar}, \bibinfo{person}{Dominika Rogozińska}, \bibinfo{person}{Dustin Herbison}, \bibinfo{person}{Elisa Bandy}, \bibinfo{person}{Emma Wang}, \bibinfo{person}{Eric Noland}, \bibinfo{person}{Erica Moreira}, \bibinfo{person}{Evan Senter},
  \bibinfo{person}{Evgenii Eltyshev}, \bibinfo{person}{Francesco Visin}, \bibinfo{person}{Gabriel Rasskin}, \bibinfo{person}{Gary Wei}, \bibinfo{person}{Glenn Cameron}, \bibinfo{person}{Gus Martins}, \bibinfo{person}{Hadi Hashemi}, \bibinfo{person}{Hanna Klimczak-Plucińska}, \bibinfo{person}{Harleen Batra}, \bibinfo{person}{Harsh Dhand}, \bibinfo{person}{Ivan Nardini}, \bibinfo{person}{Jacinda Mein}, \bibinfo{person}{Jack Zhou}, \bibinfo{person}{James Svensson}, \bibinfo{person}{Jeff Stanway}, \bibinfo{person}{Jetha Chan}, \bibinfo{person}{Jin~Peng Zhou}, \bibinfo{person}{Joana Carrasqueira}, \bibinfo{person}{Joana Iljazi}, \bibinfo{person}{Jocelyn Becker}, \bibinfo{person}{Joe Fernandez}, \bibinfo{person}{Joost van Amersfoort}, \bibinfo{person}{Josh Gordon}, \bibinfo{person}{Josh Lipschultz}, \bibinfo{person}{Josh Newlan}, \bibinfo{person}{Ju yeong Ji}, \bibinfo{person}{Kareem Mohamed}, \bibinfo{person}{Kartikeya Badola}, \bibinfo{person}{Kat Black}, \bibinfo{person}{Katie Millican}, \bibinfo{person}{Keelin
  McDonell}, \bibinfo{person}{Kelvin Nguyen}, \bibinfo{person}{Kiranbir Sodhia}, \bibinfo{person}{Kish Greene}, \bibinfo{person}{Lars~Lowe Sjoesund}, \bibinfo{person}{Lauren Usui}, \bibinfo{person}{Laurent Sifre}, \bibinfo{person}{Lena Heuermann}, \bibinfo{person}{Leticia Lago}, \bibinfo{person}{Lilly McNealus}, \bibinfo{person}{Livio~Baldini Soares}, \bibinfo{person}{Logan Kilpatrick}, \bibinfo{person}{Lucas Dixon}, \bibinfo{person}{Luciano Martins}, \bibinfo{person}{Machel Reid}, \bibinfo{person}{Manvinder Singh}, \bibinfo{person}{Mark Iverson}, \bibinfo{person}{Martin Görner}, \bibinfo{person}{Mat Velloso}, \bibinfo{person}{Mateo Wirth}, \bibinfo{person}{Matt Davidow}, \bibinfo{person}{Matt Miller}, \bibinfo{person}{Matthew Rahtz}, \bibinfo{person}{Matthew Watson}, \bibinfo{person}{Meg Risdal}, \bibinfo{person}{Mehran Kazemi}, \bibinfo{person}{Michael Moynihan}, \bibinfo{person}{Ming Zhang}, \bibinfo{person}{Minsuk Kahng}, \bibinfo{person}{Minwoo Park}, \bibinfo{person}{Mofi Rahman},
  \bibinfo{person}{Mohit Khatwani}, \bibinfo{person}{Natalie Dao}, \bibinfo{person}{Nenshad Bardoliwalla}, \bibinfo{person}{Nesh Devanathan}, \bibinfo{person}{Neta Dumai}, \bibinfo{person}{Nilay Chauhan}, \bibinfo{person}{Oscar Wahltinez}, \bibinfo{person}{Pankil Botarda}, \bibinfo{person}{Parker Barnes}, \bibinfo{person}{Paul Barham}, \bibinfo{person}{Paul Michel}, \bibinfo{person}{Pengchong Jin}, \bibinfo{person}{Petko Georgiev}, \bibinfo{person}{Phil Culliton}, \bibinfo{person}{Pradeep Kuppala}, \bibinfo{person}{Ramona Comanescu}, \bibinfo{person}{Ramona Merhej}, \bibinfo{person}{Reena Jana}, \bibinfo{person}{Reza~Ardeshir Rokni}, \bibinfo{person}{Rishabh Agarwal}, \bibinfo{person}{Ryan Mullins}, \bibinfo{person}{Samaneh Saadat}, \bibinfo{person}{Sara~Mc Carthy}, \bibinfo{person}{Sarah Cogan}, \bibinfo{person}{Sarah Perrin}, \bibinfo{person}{Sébastien M.~R. Arnold}, \bibinfo{person}{Sebastian Krause}, \bibinfo{person}{Shengyang Dai}, \bibinfo{person}{Shruti Garg}, \bibinfo{person}{Shruti Sheth},
  \bibinfo{person}{Sue Ronstrom}, \bibinfo{person}{Susan Chan}, \bibinfo{person}{Timothy Jordan}, \bibinfo{person}{Ting Yu}, \bibinfo{person}{Tom Eccles}, \bibinfo{person}{Tom Hennigan}, \bibinfo{person}{Tomas Kocisky}, \bibinfo{person}{Tulsee Doshi}, \bibinfo{person}{Vihan Jain}, \bibinfo{person}{Vikas Yadav}, \bibinfo{person}{Vilobh Meshram}, \bibinfo{person}{Vishal Dharmadhikari}, \bibinfo{person}{Warren Barkley}, \bibinfo{person}{Wei Wei}, \bibinfo{person}{Wenming Ye}, \bibinfo{person}{Woohyun Han}, \bibinfo{person}{Woosuk Kwon}, \bibinfo{person}{Xiang Xu}, \bibinfo{person}{Zhe Shen}, \bibinfo{person}{Zhitao Gong}, \bibinfo{person}{Zichuan Wei}, \bibinfo{person}{Victor Cotruta}, \bibinfo{person}{Phoebe Kirk}, \bibinfo{person}{Anand Rao}, \bibinfo{person}{Minh Giang}, \bibinfo{person}{Ludovic Peran}, \bibinfo{person}{Tris Warkentin}, \bibinfo{person}{Eli Collins}, \bibinfo{person}{Joelle Barral}, \bibinfo{person}{Zoubin Ghahramani}, \bibinfo{person}{Raia Hadsell}, \bibinfo{person}{D. Sculley},
  \bibinfo{person}{Jeanine Banks}, \bibinfo{person}{Anca Dragan}, \bibinfo{person}{Slav Petrov}, \bibinfo{person}{Oriol Vinyals}, \bibinfo{person}{Jeff Dean}, \bibinfo{person}{Demis Hassabis}, \bibinfo{person}{Koray Kavukcuoglu}, \bibinfo{person}{Clement Farabet}, \bibinfo{person}{Elena Buchatskaya}, \bibinfo{person}{Sebastian Borgeaud}, \bibinfo{person}{Noah Fiedel}, \bibinfo{person}{Armand Joulin}, \bibinfo{person}{Kathleen Kenealy}, \bibinfo{person}{Robert Dadashi}, {and} \bibinfo{person}{Alek Andreev}.} \bibinfo{year}{2024}\natexlab{}.
\newblock \bibinfo{title}{Gemma 2: Improving Open Language Models at a Practical Size}.
\newblock
\showeprint[arxiv]{2408.00118}~[cs.CL]
\urldef\tempurl%
\url{https://arxiv.org/abs/2408.00118}
\showURL{%
\tempurl}


\bibitem[Tubaro et~al\mbox{.}(2020)]%
        {tubaro2020trainerVerifierImitator}
\bibfield{author}{\bibinfo{person}{Paola Tubaro}, \bibinfo{person}{Antonio~A. Casilli}, {and} \bibinfo{person}{Marion Coville}.} \bibinfo{year}{2020}\natexlab{}.
\newblock \showarticletitle{The Trainer, the Verifier, the Imitator: Three Ways in Which Human Platform Workers Support Artificial Intelligence}.
\newblock \bibinfo{journal}{\emph{Big Data \& Society}} \bibinfo{volume}{7}, \bibinfo{number}{1} (\bibinfo{year}{2020}).
\newblock
\href{https://doi.org/10.1177/2053951720919776}{doi:\nolinkurl{10.1177/2053951720919776}}


\bibitem[{\"U}nver(2024)]%
        {unver2024aiHumanRights}
\bibfield{author}{\bibinfo{person}{H.~Akin {\"U}nver}.} \bibinfo{year}{2024}\natexlab{}.
\newblock \bibinfo{booktitle}{\emph{Artificial Intelligence and Human Rights: using {AI} as a weapon of repression and its impact on human rights}}.
\newblock \bibinfo{type}{{T}echnical {R}eport}. \bibinfo{institution}{European Parliament}.
\newblock
\urldef\tempurl%
\url{https://www.europarl.europa.eu/RegData/etudes/IDAN/2024/754450/EXPO_IDA(2024)754450_EN.pdf}
\showURL{%
\tempurl}


\bibitem[Wang et~al\mbox{.}(2024)]%
        {wang2024standGuard}
\bibfield{author}{\bibinfo{person}{Min Wang}, \bibinfo{person}{Peng Lin}, \bibinfo{person}{Siyuan Cai}, \bibinfo{person}{Shaohua An}, \bibinfo{person}{Shiyu Ma}, \bibinfo{person}{Ziming Lin}, \bibinfo{person}{Chao Huang}, {and} \bibinfo{person}{Bo Xu}.} \bibinfo{year}{2024}\natexlab{}.
\newblock \bibinfo{title}{STAND-Guard: A Small Task-Adaptive Content Moderation Model}.
\newblock \bibinfo{howpublished}{arXiv}.
\newblock
\urldef\tempurl%
\url{https://arxiv.org/abs/2411.05214}
\showURL{%
\tempurl}


\bibitem[Wang et~al\mbox{.}(2025)]%
        {wang2025probingassociationbiasesllm}
\bibfield{author}{\bibinfo{person}{Yuxin Wang}, \bibinfo{person}{Botao Yu}, \bibinfo{person}{Ivory Yang}, \bibinfo{person}{Saeed Hassanpour}, {and} \bibinfo{person}{Soroush Vosoughi}.} \bibinfo{year}{2025}\natexlab{}.
\newblock \bibinfo{title}{Probing Association Biases in LLM Moderation Over-Sensitivity}.
\newblock
\showeprint[arxiv]{2505.23914}~[cs.CL]
\urldef\tempurl%
\url{https://arxiv.org/abs/2505.23914}
\showURL{%
\tempurl}


\bibitem[Wen et~al\mbox{.}(2025)]%
        {10.1145/3726986.3727014}
\bibfield{author}{\bibinfo{person}{Ruoyu Wen}, \bibinfo{person}{Stephanie Crowe}, \bibinfo{person}{Kunal Gupta}, \bibinfo{person}{Xinyue Li}, \bibinfo{person}{Mark Billinghurst}, \bibinfo{person}{Simon Hoermann}, \bibinfo{person}{D~D Allan}, \bibinfo{person}{Alaeddin Nassani}, {and} \bibinfo{person}{Thammathip Piumsomboon}.} \bibinfo{year}{2025}\natexlab{}.
\newblock \showarticletitle{Large Language Models for Automatic Detection of Sensitive Topics}. In \bibinfo{booktitle}{\emph{Proceedings of the 36th Australasian Conference on Human-Computer Interaction}} \emph{(\bibinfo{series}{OzCHI '24})}. \bibinfo{publisher}{Association for Computing Machinery}, \bibinfo{address}{New York, NY, USA}, \bibinfo{pages}{500–514}.
\newblock
\showISBNx{9798400715099}
\href{https://doi.org/10.1145/3726986.3727014}{doi:\nolinkurl{10.1145/3726986.3727014}}


\bibitem[Willner and Chakrabarti(2024)]%
        {willner2024policyDrivenTechPolicyPress}
\bibfield{author}{\bibinfo{person}{David Willner} {and} \bibinfo{person}{Samidh Chakrabarti}.} \bibinfo{year}{2024}\natexlab{}.
\newblock \bibinfo{booktitle}{\emph{Using {LLMs} for Policy-Driven Content Classification}}.
\newblock Tech Policy Press.
\newblock
\urldef\tempurl%
\url{https://www.techpolicy.press/using-llms-for-policy-driven-content-classification/}
\showURL{%
\tempurl}


\bibitem[Yin et~al\mbox{.}(2025)]%
        {yin2025bingoguard}
\bibfield{author}{\bibinfo{person}{Fan Yin}, \bibinfo{person}{Philippe Laban}, \bibinfo{person}{Xiaocheng Peng}, \bibinfo{person}{Yichao Zhou}, \bibinfo{person}{Yuning Mao}, \bibinfo{person}{Vikhyat Vats}, \bibinfo{person}{Lee Ross}, \bibinfo{person}{Divyanshu Agarwal}, \bibinfo{person}{Caiming Xiong}, {and} \bibinfo{person}{Chao Wu}.} \bibinfo{year}{2025}\natexlab{}.
\newblock \bibinfo{title}{BingoGuard: LLM Content Moderation Tools with Risk Levels}.
\newblock \bibinfo{howpublished}{arXiv}.
\newblock
\urldef\tempurl%
\url{https://arxiv.org/abs/2503.06550}
\showURL{%
\tempurl}


\bibitem[Zeng et~al\mbox{.}(2024)]%
        {zeng2024shieldgemma}
\bibfield{author}{\bibinfo{person}{Wenhui Zeng}, \bibinfo{person}{Yifan Liu}, \bibinfo{person}{Ryan Mullins}, \bibinfo{person}{Logan Peran}, \bibinfo{person}{Jose Fernandez}, \bibinfo{person}{Hamza Harkous}, \bibinfo{person}{Karthik Narasimhan}, \bibinfo{person}{Drew Proud}, \bibinfo{person}{Pratyush Kumar}, \bibinfo{person}{Bhargav Radharapu}, \bibinfo{person}{Orion Sturman}, {and} \bibinfo{person}{Oscar Wahltinez}.} \bibinfo{year}{2024}\natexlab{}.
\newblock \bibinfo{title}{ShieldGemma: Generative {AI} content moderation based on Gemma}.
\newblock \bibinfo{howpublished}{arXiv}.
\newblock
\urldef\tempurl%
\url{https://arxiv.org/abs/2407.21772}
\showURL{%
\tempurl}


\end{thebibliography}
\end{document}